\documentclass[letterpaper]{article} 
\usepackage{aaai2026}  
\usepackage{times}  
\usepackage{helvet}  
\usepackage{courier}  
\usepackage[hyphens]{url}  
\usepackage{graphicx} 
\usepackage{natbib}  
\usepackage{caption} 
\usepackage{algorithm}
\usepackage{algorithmic}
\usepackage{amsmath}
\usepackage{amsfonts} 
\usepackage{diagbox}
\usepackage{booktabs}
\usepackage{multirow}
\usepackage{longtable}

\usepackage{newfloat}
\usepackage{listings}
\DeclareCaptionStyle{ruled}{labelfont=normalfont,labelsep=colon,strut=off} 
\floatstyle{ruled}
\newfloat{listing}{tb}{lst}{}
\floatname{listing}{Listing}
\title{Guideline‑Consistent Segmentation via Multi‑Agent Refinement}
\author{
    Vanshika Vats,
    Ashwani Rathee,
    James Davis
}
\affiliations{
    University of California, Santa Cruz\\
    1156 High St, Santa Cruz, CA 95064

    \{vvats, arathee1, davisje\}@ucsc.edu
}

\usepackage{bibentry}

\begin{document}

\maketitle

\begin{abstract}
Semantic segmentation in real-world applications often requires not only accurate masks but also strict adherence to textual labeling guidelines. These guidelines are typically complex and long, and both human and automated labeling often fail to follow them faithfully. Traditional approaches depend on expensive task-specific retraining that must be repeated as the guidelines evolve. Although recent open-vocabulary segmentation methods excel with simple prompts, they often fail when confronted with sets of paragraph-length guidelines that specify intricate segmentation rules. To address this, we introduce a multi-agent, training-free framework that coordinates general-purpose vision-language models within an iterative Worker-Supervisor refinement architecture. The \textit{Worker} performs the segmentation, the \textit{Supervisor} critiques it against the retrieved guidelines, and a lightweight reinforcement learning stop policy decides when to terminate the loop, ensuring guideline-consistent masks while balancing resource use. Evaluated on the Waymo and ReasonSeg datasets, our method notably outperforms state-of-the-art baselines, demonstrating strong generalization and instruction adherence. 
\end{abstract}

\begin{links}
    \link{Project page}{https://guideline-seg.github.io/}
\end{links}

\section{Introduction}
\label{sec:intro}

Pixel-accurate semantic image segmentation is crucial for safety-critical domains such as autonomous driving, robotics, and medical imaging \cite{elhassan2024realtimeseg_ad_survey, al2024review, wang2022medical}. Traditionally, segmentation models have relied heavily on supervised training on pre-defined classes that generalize poorly to novel categories, requiring costly re-annotation and retraining whenever new classes emerge \cite{ronneberger2015u, chen2018encoder}. Recent advancements have introduced open-vocabulary semantic segmentation, leveraging large-scale Vision-Language Models (VLMs) to recognize arbitrary object categories via textual prompts \cite{liang2023open,ren2024grounded}.

\begin{figure}[ht!]
  \centering
  \includegraphics[width=\linewidth]{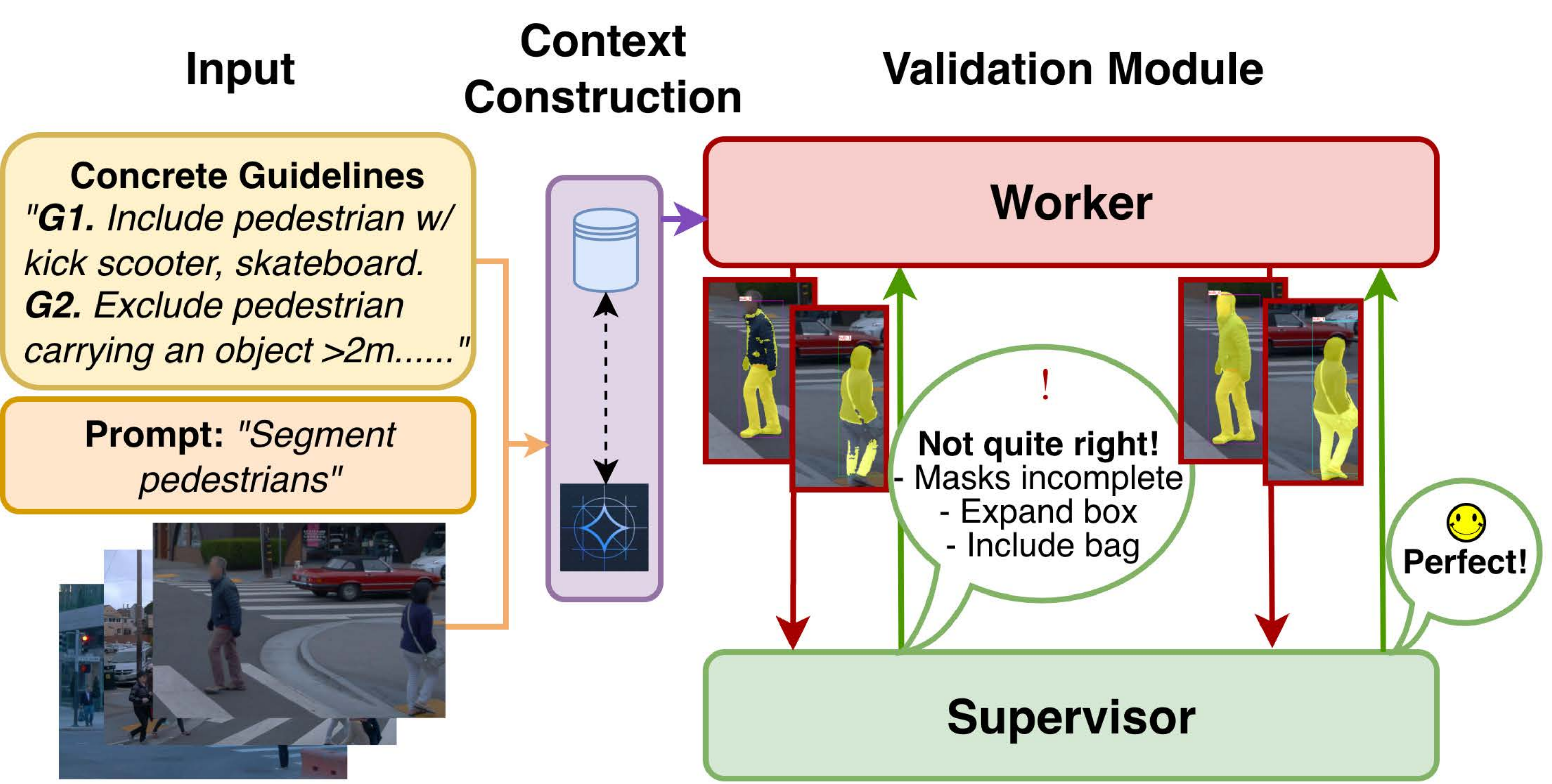}
  \caption{Method overview of our Worker–Supervisor iterative loop for guideline-consistent semantic segmentation.}
  \label{fig:small_overview}
\end{figure}

\begin{figure}[ht!]
  \centering
  \includegraphics[width=\linewidth]{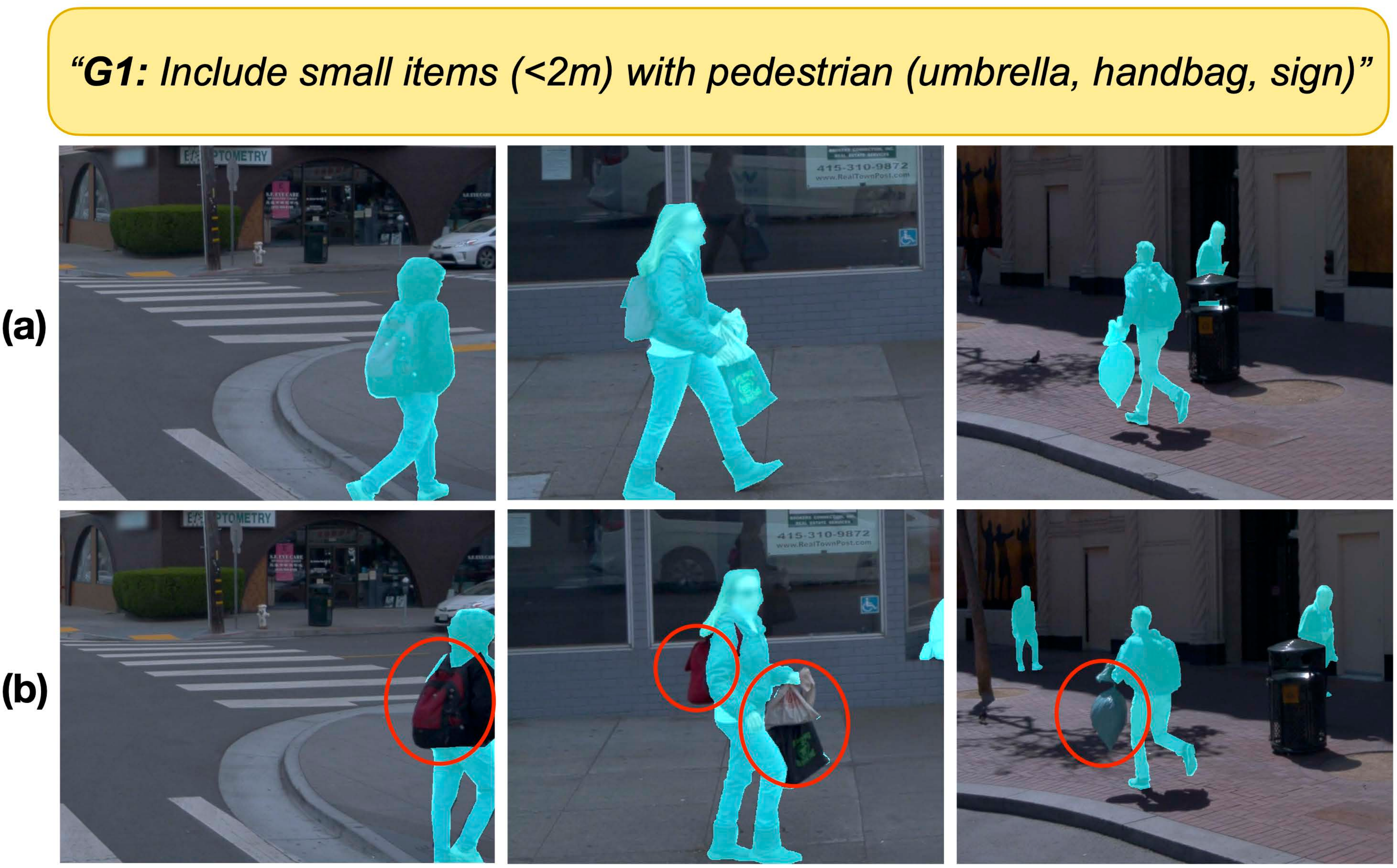}
  \caption{Annotation inconsistencies in the Waymo segmentation dataset. Row (a) shows masks adhere to guideline G1 at timestamp $t$, while row (b), from the same scene at $t+\Delta t$, shows a clear violation of the same guideline.}
  \label{fig:waymo_inconsistent}
\end{figure}

Semantic segmentation is not merely about generating accurate masks, but producing the \textit{correct} masks that align with the intended labeling policy, \textit{every time}. 
These decisions often rely on strict, text-defined rules. For instance, Waymo dataset \cite{waymo} specifies that \textit{pedestrian} class must \textit{`include skateboard riders, exclude mannequins'}. Similarly, in land‑cover segmentation  \cite{Demir2018DeepGlobe2A}, the class \textit{range land} should cover \textit{`Any non-forest, non-farm, green land, grass'}. When models ignore these intricate details, even pixel‑accurate masks can be functionally wrong.

\begin{figure*}[ht]
  \centering
  \includegraphics[width=\linewidth]{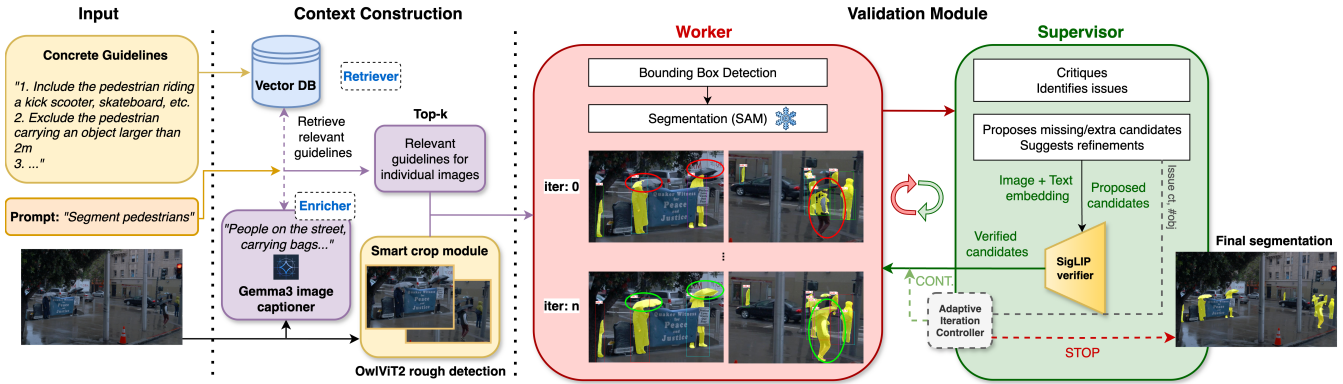}
  \caption{Overview of our pipeline. The input image and textual guidelines undergo context construction to extract scene-relevant rules. This focused context is processed through an iterative VLM loop where the \textit{Worker} segments and the \textit{Supervisor} critiques and suggests improvements. The number of iterations is controlled by an adaptive iteration controller. The final output is a segmentation that faithfully adheres to long and detailed guidelines.}
  \label{fig:full_pipeline}
\end{figure*}

\noindent Existing open‑vocabulary systems excel on short prompts \cite{lisa, sun2025visual} but crumble under long intricate rules. Worse, some public `ground‑truth' masks produced by costly human-annotators, themselves often violate their own manuals. As shown in Fig.~\ref{fig:waymo_inconsistent}, the same scene at different timestamps in the Waymo dataset exhibits inconsistent masks that ignore the guidelines. Domain‑specific detectors and fine‑tuned models might be adapted for guideline‑aware segmentation, but only after costly retraining. Labeling rules can evolve, and relying on continuous retraining is impractical. These inconsistencies reveal the difficulty of the task of rule-adhering segmentation and undermine the reliability of conventional benchmarks. 

In this study, we achieve guideline-consistent semantic segmentation via multi-agent, training-free framework. Leveraging VLMs (Gemini-2.5-flash-preview \cite{comanici2025gemini}) as agents, our approach employs a hierarchical Worker-Supervisor system where each agent plays a distinct role, iteratively generating, critiquing, refining, and validating segmentation outputs (Fig. \ref{fig:small_overview}). Specifically, the \textit{Worker} module first produces initial detections based on enriched image-specific contexts. These detections are then segmented by a frozen Segment Anything Model (SAM) \cite{kirillov2023segany, ravi2024sam2}. Subsequently, the \textit{Supervisor} module evaluates the Worker's segmentations against the given guidelines, identifies errors, and proposes refinements. This loop continues until a reinforcement learning-based stop criterion terminates the process or a maximum iteration count is reached. Crucially, both the VLM and SAM remain frozen; all adaptation occurs through strategic prompting and context-aware control, making our method uniquely flexible, scalable, and capable of handling complex, paragraph-length instructions.

Our method outperforms existing state-of-the-art (SOTA) approaches, delivering a notable increase in guideline-consistent segmentation performance on the Waymo and ReasonSeg datasets. This highlights our framework's superior ability to produce precise segmentations that closely adhere to intricate, lengthy guidelines without requiring model retraining. Thus, the main contribution of our work is a novel framework for training-free guideline-consistent semantic segmentation, capable of interpreting and adhering to long and detailed labeling instructions.

\section{Related Works}
\subsubsection{Vision‑Language Models for Segmentation.}
Vision-Language models (VLMs) align visual features with language \cite{radford2021learning,li2022blip,liu2023llava} and are increasingly being used in object detection \cite{liu2024grounding, Cheng2024YOLOWorld} and segmentation \cite{liang2023open, xia2024gsva}. Recently, promptable segmentation has emerged as a flexible paradigm where an image and user prompts jointly determine the output mask. The prompts are typically either points/boxes \cite{kirillov2023segany} or text-conditioned \cite{ren2024grounded}. Unified VL segmenters \cite{lisa, seem, sun2025visual} embed multi-modal prompts into a single transformer, enabling open‑set and multi‑prompt segmentation. Reasoning-based segmenters further combine language reasoning with segmentation, either through chain-of-thought style guidance or specialized \texttt{<seg>} tokens that steer mask prediction \cite{segzero,read}. However, these methods typically handle only single words or short sentences, often performing poorly on longer contexts. More recent models like Gemini-2.5 \cite{comanici2025gemini}, Qwen2.5‑VL \cite{Qwen2.5-VL}, and GPT‑4V \cite{openai2024gpt4v} offer improved comprehension of longer textual inputs but still fall short of accurate detection and segmentation. Moreover, when operating in a single pass, they offer limited understanding of lengthy intricate rules, complex constraints, or structured verification guideline compliance. Using Gemini‑2.5 as our baseline, we address these concerns by introducing an iterative‑refinement and verification framework that yields guideline‑consistent segmentation. This choice was driven by the need to effectively process longer contexts, a capability not yet reliably offered by current open-weight models.

\subsubsection{Agents and Self‑Correction.}
Recent research frames large models as smaller, specialized agents that collaborate to complete complex tasks \cite{wu2023visual, shen2023hugginggpt}. These agents iteratively refine their outputs based on mutual feedback without additional training \cite{self_refine, reflexion}. In vision, VLMs leverage self‑correction for iterative re‑grounding \cite{Liao2025SelfCorrection, Su2024ScanFormer, Sun2021IterativeShrinking}, training‑free contrastive guidance \cite{Wan2024CRG}, and visual prompting \cite{yang2023set}. Only a few agent-based approaches have addressed semantic segmentation \cite{sun2025visual}, and existing ones neither produce accurate masks under long instruction nor include explicit self-critique mechanisms. To bridge these gaps, we propose a Worker-Supervisor agent framework that iteratively refines the segmentations, effectively addressing missed detections and pruning false positives under complex guidelines. Such agent-based workflows are well-suited to semantic segmentation tasks because they inherently require multiple steps: segmenting, verifying, refining, and adaptively determining when to stop, and thus greatly benefit from iterative planning and structured feedback under detailed, rule-rich instructions.

\section{Method}
\label{sec:method}

Fig.~\ref{fig:full_pipeline} summarizes our pipeline. Given the input image and a set of guidelines, we first build the context from the image description and retrieve relevant guidelines, followed by a loop of Worker → Supervisor → Worker ... till the Adaptive Iteration Controller (AiRC) decides when to stop or upon reaching a maximum number of predecided iterations.

\subsection{Inputs}
We begin with an input image $Img_i$ that must be segmented based on a prompt $P$, while strictly following a set of detailed guidelines. For instance, in the Waymo dataset \cite{waymo}, inclusion rules may state \textit{`Include a small child or small items (e.g., umbrella, handbag, sign) with the pedestrian'}, and exclusion rules may require \textit{`Exclude mannequins, statues, dummies, covered objects, billboards, posters, indoor pedestrians, and reflections'}. To enable structured interpretation, we use GPT-4o \cite{hurst2024gpt} to convert them into a structured JSON format, assigning each a unique ID $G = \{G_0, G_1, \dots, G_m\}$ and a brief description. This facilitates the context construction.

\begin{figure*}[ht!]
  \centering
  \includegraphics[width=\linewidth]{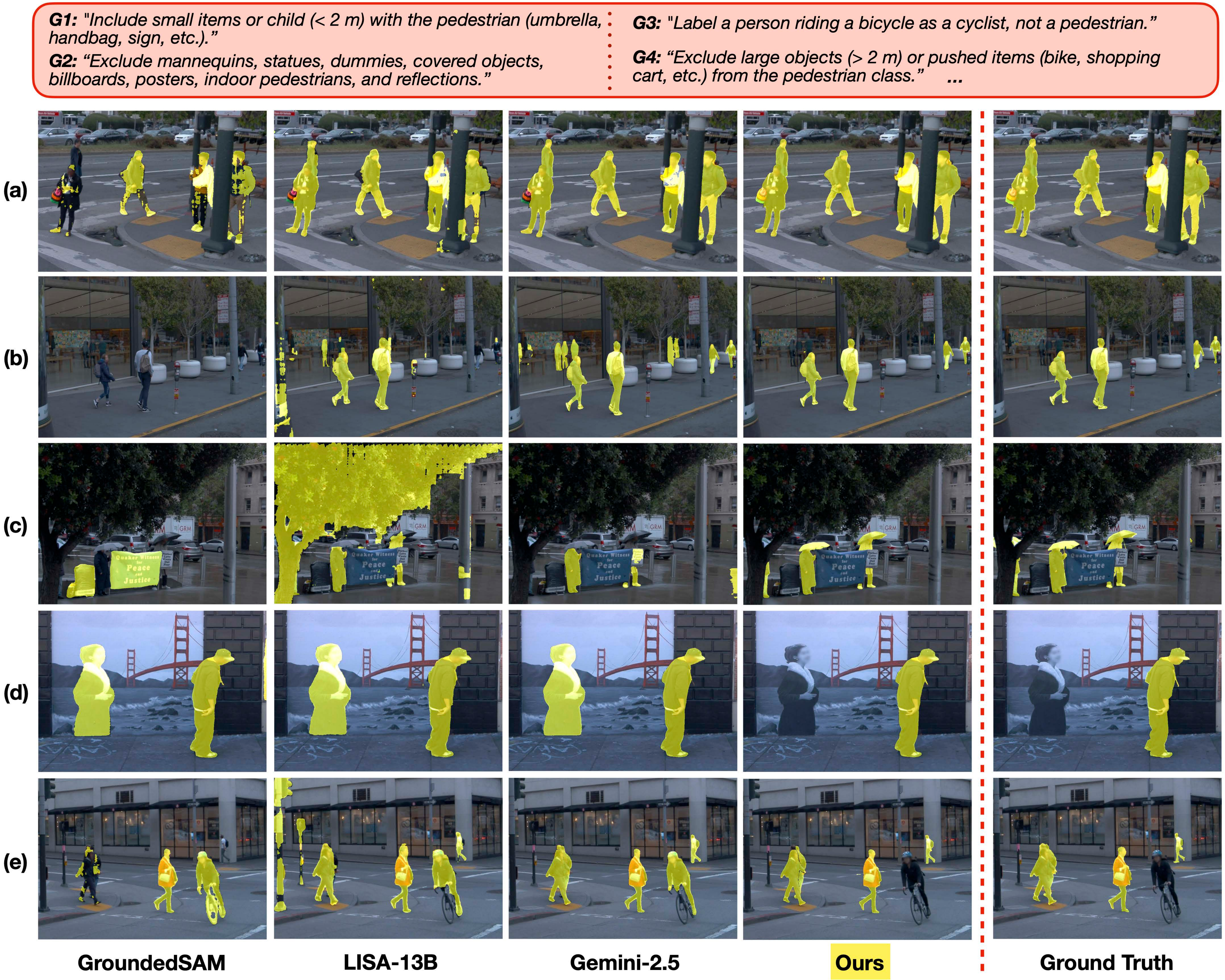}
  \caption{Qualitative comparison of our method with other state-of-the-art. Our method successfully complies with guidelines G1 [rows-(a),(c)], G2 [rows-(b),(d)], G3 and G4 [row-(e)]. More examples in supplementary material.}
  \label{fig:comparison}
\end{figure*}


\subsection{Context Construction}
\label{sec:context}
Each image is different, and not all guidelines are relevant to every scene. For example, for an image without bicycles or cyclists, \textit{`G3: A person riding a bicycle is not labeled as a pedestrian, but labeled as a cyclist'} becomes irrelevant. Including such unnecessary instructions can overwhelm the pipeline with irrelevant details. Thus, we construct image-specific contexts by retrieving only the most relevant guidelines.

We encode all textual guidelines into a FAISS \cite{johnson2019billion} vector database (DB), enabling fast similarity-based retrieval. Storing guidelines in a vector DB enables quick filtering of the most relevant rules for a given image. This follows a context construction through an \textit{Enricher}, which describes the scene and enriches the retrieval query, and a \textit{Retriever}, which searches for the most relevant guidelines from the vector DB. All context modules are lightweight and add minimal computational overhead.

\subsubsection{Enricher.}
To retrieve the most relevant guidelines for a given image, we first generate a concise yet informative textual description of the scene. Gemma3-4B \cite{team2025gemma}, a lightweight multimodal model, is used to caption the image based on the prompt. We then form a text-rich query $Q = \{P,\ {<caption>},\ H \times W\}$, where $H \times W$ represents the image resolution. Including resolution helps account for scale-dependent rules like, ‘Ignore distant pedestrians’, ‘Exclude oversized objects’, enriching the semantic context. This query is encoded into a vector ($Q_v$) using the SentenceTransformer \cite{reimers-2020-multilingual-sentence-bert} (all-MiniLM-L6-v2) to semantically fetch relevant guidelines from the FAISS DB.

\subsubsection{Retriever.}
The Retriever module uses $Q_v$ to perform a cosine similarity search over the FAISS index, retrieving the top-$k$ most relevant guidelines $G_k\subset G$. This selective retrieval avoids overloading the downstream Validation module with unnecessary rules, ensuring the context remains concise, focused, and aligned with the image content.

\subsubsection{Smart Crop.}
Modern VLMs often struggle with small or distant objects due to limited attention resolution. To mitigate this, we employ a smart cropping strategy that divides the image into two focused regions based on rough object locations. We first downscale the image $Img$ to $0.8\times$ and pass it through OWLv2 \cite{minderer2023scaling} to obtain rough, noisy bounding boxes ($B$) for the target class. This gives us focused regions of interest.

To avoid splitting objects and maintain balance, the module analyzes object distribution and spatial layout. It chooses a vertical split that keeps object counts roughly equal on both sides and maximizes the gap between clusters, ensuring clean separation. The resulting crops $x_\ell$ and $x_r$, defined by $\max_x(B_{\text{left}})$ and $\min_x(B_{\text{right}})$, preserve semantic coherence and are passed to the Validation module for processing.

\subsection{Validation Module}
The Validation module handles segmentation through an iterative loop composed of two primary components: the \textbf{Worker}, which performs object detection and segmentation, and the \textbf{Supervisor}, which evaluates the Worker’s output against the retrieved guidelines and proposes corrections. An \textbf{Adaptive Iteration Controller (AiRC)} governs how many refinement rounds are performed. AiRC monitors factors such as object count and unresolved issues to decide whether to continue or stop the loop, up to a maximum iteration limit.

\subsubsection{Worker.}
The Worker module is dedicated to executing the segmentation task and comprises two stages: an initial detection step and a refinement step across $n$ iterations, incorporating Supervisor feedback.

At initialization, the Worker identifies bounding boxes for the target class based on the user prompt $P$, returning coordinates in the form $[y_{min}, x_{min}, y_{max}, x_{max}]$, class labels, and a unique ID $Subjects:\{sub_0, sub_1, \dots, sub_m\}$ to each subject. These are passed to a frozen SAM model to produce segmentation masks. Note that this SAM model is not fine-tuned or retrained on any particular dataset to maintain its generalizability; we rely on carefully crafted prompt inputs (boxes, points) to guide it in producing accurate masks. 

Since general-purpose VLMs are not optimized for coordinate-precise localization, the initial boxes are often coarse or misaligned, potentially leading to incorrect or incomplete masks. This inaccuracy propagates to the SAM segmentation outputs. Moreover, SAM, while powerful, is task-agnostic and may overlook rule-specific inclusion/exclusion criteria. To correct such issues, the Worker triggers the Supervisor, which audits the output and provides corrective feedback for iterative refinement.

\subsubsection{Supervisor.}
The Supervisor is responsible for evaluating and critiquing the outputs generated by the Worker. Following the principle that a group of smaller, specialized agents often performs better than a single agent overloaded with responsibilities \cite{hong2023metagpt, wu2024autogen}, we divide the Supervisor into two distinct components:

$Agent\ 1$-\texttt{Supervisor\_eval}: This agent has access to the image, the Worker’s output, and the relevant guidelines $G_k$. It identifies three types of issues: (i) missing objects that should have been segmented, (ii) false positives that violate exclusion rules, and (iii) refinement opportunities for imperfect masks. Each issue is recorded in a structured JSON, specifying object type, brief justification, and precise refinement suggestions such as \textit{`expand box to the right to include hand'} or \textit{`shrink box to exclude the background'}. These are passed to Agent 2 for candidate box generation.

$Agent\ 2$-\texttt{Supervisor\_boxgen}:  This agent receives the missing\_objects, false\_positives, and corresponding critiques from Agent 1, and generates bounding boxes for each proposed candidate. However, since these items are often small and lack rich visual or textual context, the proposed boxes tend to be imprecise, undersized, or occasionally misplaced. We, thus, employ a SigLIP-based verifier.

\subsubsection{SigLIP Verifier.}
To verify candidates from Agent 2, we crop image regions with a small buffer to retain context (e.g., a tightly cropped bag alone may be unclear, but showing it being held helps in recognition). These image crops are passed through a SigLIP \cite{zhai2023sigmoid} image encoder, while the associated labels are encoded using a text encoder. SigLIP outputs a logit for each image-text pair, which we convert to a probability using a sigmoid function. If this probability exceeds a threshold, the candidate is considered a valid suggestion by the Supervisor; otherwise, it is discarded.

The verified candidates and refinement instructions are returned to the Worker for final adjustments. For missing objects, the Worker accepts verified bounding boxes and uses SAM to generate the corresponding masks. False positives are removed by prompting SAM with the identified box and a negative point label to erase the mask.  For refinements, the Worker modifies existing bounding boxes based on the Supervisor's suggestions before re-generating the masks.

\subsection{Adaptive Iteration Controller}
Each iteration in our loop requires three API calls, so the overall cost scales with the number of iterations. A fixed count could be used, but it would not adapt well across varying scenes, i.e., too many iterations may waste resources, while too few might cut off meaningful improvements. To address this, we implement a reinforcement learning (RL) policy that dynamically determines when to stop the loop. This adaptive iteration controller (AiRC) tracks scene density and unresolved guideline violations, and the policy selects \textsc{CONTINUE} or \textsc{STOP} accordingly. We frame the iteration‑control problem as a finite‑horizon Markov Decision Process (MDP) with the following elements.

\subsubsection{Issue count.} 
We quantify Supervisor's feedback as issue count $I$, which shapes the reward function. Each genuine guideline error (missed or false object) counts 1 point, driving the learning signal. Refinements usually suggest low-impact mask tweaks and can be repetitive and noisy, and thus count as 0.1 points.
Thus, the issue count $I_{t}$ at the end of each iteration becomes
\begin{equation}
    I_{t} = I_{miss} + I_{false} + 0.1I_{ref}
\end{equation}

\subsubsection{State space.}
The state representation captures both the scene complexity and the presence of unresolved guideline-violating issues $I$. Each $Img$ crop is mapped to one of six abstract states, defined as $ s = 2d + v$
where $d \in \{0,1,2\}$ is a coarse scene‑complexity density bucket (few/medium/crowd) computed from the initial object count in the crop, and $v \in \{0,1\}$ flags whether residual guideline violations remain after the current pass (clean=0, dirty=1).

\subsubsection{Actions.} 
There exist two possible actions at each step: either to stop the iterative process or proceed to the next step, up to a predefined   \texttt{MAX\_ITERS}.
$a\in\{\text{STOP}, \text{CONTINUE}\}$.

\subsubsection{Immediate reward.}
The immediate reward reflects the change in issue count $I_{t}$ across iterations. At iteration $t$:
\begin{equation}
r(s,a,s') = 
\begin{cases}
\begin{aligned}
\underbrace{(I_t - I_{t+1})}_{\text{issues fixed}}
-\,\underbrace{c}_{\text{step cost}} \\
\quad+\;\underbrace{b\,[I_{t+1}=0]}_{\text{early-resolve bonus}},
\end{aligned}
&\quad a_t = \mathrm{CONTINUE},\\[8pt]
0, 
&\quad a_t = \mathrm{STOP},I_t=0,\\[4pt]
-\underbrace{p}_{\text{early-stop penalty}}, 
&\quad a_t = \mathrm{STOP},I_t>0.
\end{cases}
\end{equation}
The reward $r$ is defined by the drop in issue count between consecutive steps, with a small step cost $c$ per iteration. This encourages the AiRC to continue only when further refinement is worthwhile. An early-stop penalty $p$ is applied if \textsc{STOP} is chosen with unresolved issues, while a clean-scene bonus $b$ is given if all issues are resolved before reaching \texttt{MAX\_ITERS}. Action values are updated with standard tabular Q‑learning \cite{RL}. To put simply, AiRC chooses \textsc{CONTINUE} if 
\begin{equation}
    Q(s_t,\mathrm{CONTINUE}) > Q(s_t,\mathrm{STOP})
\end{equation}
The module selects CONTINUE whenever the Q-value (expected discounted return) for CONTINUE in the current state exceeds the Q-value for STOP; otherwise, it stops. (More in supplementary)


We use RL instead of a fixed heuristic because the value of an additional refinement emerges only after that pass, creating a delayed, sequential reward structure that RL’s temporal‑difference updates naturally handle. RL learns exactly when the expected error reduction justifies the per‑iteration cost, adapting to each crop’s complexity without manual thresholds. In contrast, static rules like a fixed pass count or issue‑count cutoff either leave errors in dense scenes or waste effort on simple ones. A compact six‑state Q‑table delivers this adaptability with minimal overhead.

\begin{table*}[]
\renewcommand{\arraystretch}{1.8}
\centering
\resizebox{\linewidth}{!}{%
\begin{tabular}{c|ccc|ccc|ccc|ccc|ccc|ccc|c}
\hline
\multirow{2}{*}{\diagbox[width=1.7cm]{Metric}{Method}} &
  \multicolumn{3}{c|}{\begin{tabular}[c]{@{}c@{}}LISA-7B\\ \cite{lisa}\end{tabular}} &
  \multicolumn{3}{c|}{\begin{tabular}[c]{@{}c@{}}LISA-13B\\ \cite{lisa}\end{tabular}} &
  \multicolumn{3}{c|}{\begin{tabular}[c]{@{}c@{}}GroundedSAM\\ \cite{ren2024grounded}\end{tabular}} &
  \multicolumn{3}{c|}{\begin{tabular}[c]{@{}c@{}}READ\\ \cite{read}\end{tabular}} &
  \multicolumn{3}{c|}{\begin{tabular}[c]{@{}c@{}}Gemini-2.5\\ \cite{comanici2025gemini}\end{tabular}} &
  \multicolumn{3}{c|}{\begin{tabular}[c]{@{}c@{}}SegZero\\ \cite{segzero}\end{tabular}} &
  \textbf{Ours} \\ \cline{2-20} 
      & S     & C     & F     & S     & C     & F     & S     & C     & F     & S     & C     & F     & S     & C     & F     & S     & C     & F     & \textbf{F}     \\ \hline
gIOU   & 43.42 & 36.47 & 23.41 & 66.74 & 40.96 & 20.16 & 75.00 & 20.84 & 20.43 & 50.90 & 47.79 & 43.94 & 67.84 & 71.39 & 69.02 & 73.49 & 73.72 & 71.96 & \textbf{80.57} \\
cIOU   & 42.73 & 34.53 & 18.32 & 73.21 & 34.04 & 14.21 & 79.78 & 20.91 & 19.36 & 53.73 & 54.16 & 45.55 & 72.01 & 76.38 & 74.24 & 75.98 & 76.91 & 73.72 & \textbf{86.70} \\
mPr    & 85.11 & 40.27 & 24.38 & 80.77 & 46.62 & 22.41 & 88.12 & 33.16 & 29.35 & 75.63 & 67.57 & 51.74 & 79.05 & 84.18 & 80.91 & 89.80 & 88.38 & 86.34 & \textbf{91.06} \\
mRec   & 45.55 & 74.86 & 79.79 & 77.52 & 74.91 & 62.41 & 81.51 & 32.52 & 28.11 & 60.82 & 55.55 & 69.75 & 79.75 & 80.78 & 79.89 & 78.46 & 77.91 & 78.45 & \textbf{84.78} \\
mDice  & 57.14 & 36.47 & 33.17 & 77.92 & 51.60 & 28.30 & 83.90 & 27.61 & 25.63 & 63.86 & 58.19 & 59.41 & 77.75 & 80.76 & 78.75 & 83.75 & 82.15 & 80.88 & \textbf{87.20} \\ \hline
\end{tabular}%
}
\caption{Segmentation results on the Waymo \textit{guideline‑consistent} dataset. As the input guideline length increases from single-word (S) to condensed (C) to full-length (F), the performance of other SOTA methods declines, with the exception of Gemini‑2.5. Our method consistently outperforms all methods, even with full-length guidelines.}
\label{tab:waymo_new}
\end{table*}

\begin{table}[]
\scriptsize
\centering
\resizebox{0.95\columnwidth}{!}{%
\begin{tabular}{l|cc}
\hline
Method                  & gIOU & cIOU \\ \hline
SEEM \cite{seem}                   & 21.2 & 25.5 \\
GroundedSAM \cite{ren2024grounded}            & 26.0   & 14.5 \\
LISA-7B \cite{lisa}                & 46.0   & 44.4 \\
FaST \cite{sun2025visual}                    & 48.7 & 47.6 \\
LISA-7B(ft) \cite{lisa}            & 52.9 & 54.0   \\
Gemini-2.5 \cite{comanici2025gemini}             & 55.5 & 44.9 \\
LISA-13B-LLava-1.5(ft) \cite{lisa} & 57.7 & 60.3 \\
READ \cite{read}                   & 59.8 & \textbf{67.6} \\
SegZero \cite{segzero}                & 62.6 & 62.0   \\ \hline
Ours                    & \textbf{68.1} & 66.4 \\ \hline
\end{tabular}%
}
\caption{Segmentation results on the ReasonSeg \textit{val} dataset. Our method outperforms the other SOTA gIOU by a good margin.}
\label{tab:reasonseg}
\end{table}



\section{Experiments}
\label{sec:exp}

\subsection{Datasets}
We evaluate our method on two datasets, the Waymo Perception dataset \cite{waymo} and ReasonSeg \cite{lisa}. For Waymo, we focus on the \textit{Pedestrian} class, which poses a greater challenge due to its subtle object boundaries and frequent segmentation errors. In contrast, larger objects like \textit{Car} can inflate performance metrics and shadow fine-grained issues. Additionally, as noted in Fig. \ref{fig:waymo_inconsistent}, the Waymo ground truth often deviates from its own labeling guidelines. Since our method explicitly aims to enforce guideline consistency, relying on such noisy annotations could lead to misleading results. To address this, we manually curate a set of 101 samples that adhere strictly to the defined guidelines. Creating this smaller, high-quality set was a necessary and labor-intensive step to enable a fair and precise evaluation of our core contribution. The data was randomly selected from 55 diverse scenes, spanning various conditions such as sunny, rainy, and nighttime settings, ensuring broad coverage and variability in the evaluation (See supplementary). 

To further evaluate the generalizability of our method, we also test it on the ReasonSeg validation set, which contains 200 challenging in-the-wild images annotated with pixel-level masks and textual rationales that span from concise phrases to more complex reasoning prompts. While the dataset does not include full-length guidelines, its intricate queries simulate guideline-driven segmentation scenarios and support the study of explainable segmentation.

\subsection{Implementation}
Given an image $I$, prompt $P$, and guidelines $G$, we embed $G$ and index with \textsc{FAISS} \cite{johnson2019billion}. A lightweight captioner (Gemma3‑4B) \cite{team2025gemma} forms the query $Q{=}\{P,<caption>,H{\times}W\}$ to retrieve the top-$k{=}8$ rules. We resize $I$ to $0.8{\times}$ and obtain coarse boxes with OWLv2 to seed Smart‑Crop.

We employ \texttt{Gemini-2.5-flash-preview-05-20} API for both the Worker and Supervisor VLMs, using it as our baseline. The Worker runs at temperature $T=0.5$, as recommended for object detection flexibility. \texttt{Supervisor\_eval} critiques with $T=0.3$ for more deterministic reasoning, using dynamic thinking mode for improved contextual judgment. \texttt{Supervisor\_boxgen} has $T=0.5$ for bounding box suggestions. Masks are produced by \texttt{SAM2.1\_hiera\_large}. Supervisor proposals are validated with SigLIP; crops are accepted when sigmoid is $\ge 0.5$. 

To account for VLM non-determinism, we allow \texttt{MIN\_ITERS}=2 and \texttt{MAX\_ITERS}=4, with early stopping controlled by AiRC. The stop policy is a $3\times2$ tabular Q‑learner ($\alpha=0.3,\gamma=0.9$) with fixed $\varepsilon$‑greedy exploration ($\varepsilon=0.02$). The reward uses step cost $c=(-)0.02$, early‑stop penalty $p=(-)2.0$, and clean‑scene bonus $b=+1.0$. The Q‑table is persisted across runs. The inference is performed on a single RTX3080 GPU, but has the flexibility to be split across 3$\times$GPUs for efficiency.  
To validate the robustness of our framework against VLM non-determinism, we perform three independent runs with different random seeds and report means and standard deviation.

\subsection{Evaluation Metrics}
We use global intersection over union (gIOU) and complete IoU (cIOU) as primary metrics \cite{kazemzadeh-etal-2014-referitgame, giou_ciou, lisa}. gIOU is the per‑image IoU averaged over the dataset; cIOU is the ratio of the cumulative intersection to the cumulative union across all images. Because cIOU disproportionately weighs large objects more, gIOU offers a more balanced assessment. We also report mean Precision (mPr), mean Recall (mRec), and mean Dice/F1 (mDice), computed per image and averaged, to quantify false positives, false negatives, and their trade‑off.

\subsection{Results and Discussion}
Figure~\ref{fig:comparison} and Tables~\ref{tab:waymo_new}–\ref{tab:reasonseg} summarize our main results. In Table~\ref{tab:waymo_new}, we evaluate the Waymo \textit{guideline‑consistent} set under three text settings: (i) the full Waymo guidelines, (ii) a condensed set of short, crisp phrases of guidelines, and (iii) a single‑word main class name \textit{`Pedestrian'}. Since phrase-conditioned SOTA methods like LISA and GroundedSAM are limited to handling short phrases and sentences, we expect performance to drop as the referring text length grows, a trend confirmed in Table \ref{tab:waymo_new}. An exception is seen in Gemini‑2.5 having enhanced long‑context understanding and achieves better results when given the full set of guidelines. Adding our image‑specific guideline retrieval with iterative refinement yields +57.16 gIoU over the strongest phrase‑conditioned full‑guideline baseline, +11.55 gIoU over the baseline Gemini-2.5, and +8.61 gIOU (80.57$\pm$0.6) over SOTA. Results are stable over three runs: high recall (84.78$\pm$0.9) and precision (91.06$\pm$0.5) show that our method recovers missed items (e.g., umbrella, backpack) and removes false positives (e.g., cyclist, reflections) that competing methods overlook.

On the ReasonSeg dataset (Table \ref{tab:reasonseg}), we outperform the strongest baseline (SegZero) by +5.5 gIoU. This demonstrates generalization to in‑the‑wild scenes with rationale‑style instructions.


\begin{table}[t!]
\renewcommand{\arraystretch}{1.3}
\centering
\resizebox{\columnwidth}{!}{%
\begin{tabular}{c|ccc|cccc}
\hline
Dataset                    & W    & C     & S     & gIOU           & cIOU           & mPr            & mRec           \\ \hline
\multirow{3}{*}{Waymo}     & \checkmark & - & - & 69.02          & 74.24          & 80.91          & 79.89          \\
                           & \checkmark & \checkmark  & - & 73.87          & 78.40          & 88.36          & 80.81          \\
                           & \checkmark & \checkmark  & \checkmark  & \textbf{80.57} & \textbf{86.70} & \textbf{91.06}  & \textbf{84.78} \\ \hline
\multirow{2}{*}{ReasonSeg} & \checkmark & -     & - & 55.56          & 44.98          & 70.66          & 61.78          \\
                           & \checkmark & -     & \checkmark  & \textbf{68.12} & \textbf{66.35} & \textbf{77.82} & \textbf{76.06} \\ \hline
\end{tabular}%
}
\caption{Ablation table showing the benefit of iterative critiquing. W-Worker, C- Context construction, S-Supervisor.}
\label{tab:ablation}
\end{table}

\begin{figure}[t!]
  \centering
  \includegraphics[width=\columnwidth]{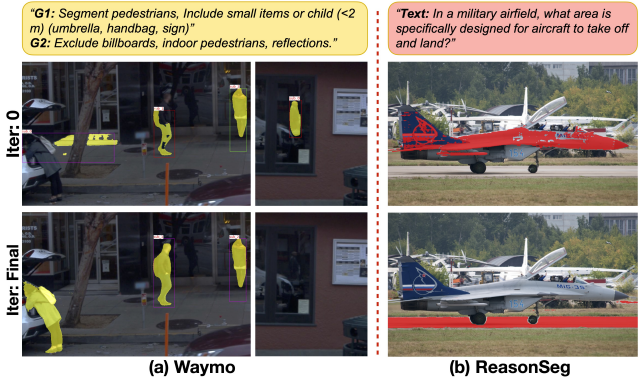}
  \caption{Examples showing how our iterative loop produces guideline-adhering segmentations over just relying on the VLM initial outputs.}
  \label{fig:ablation}
\end{figure}

Our results follow directly from the design. First, the context construction via image‑specific guideline retrieval prevents instruction overload, so the VLM reasons over only the rules relevant to the scene, explaining robustness on long guidelines. Smart crop improves initial detections of small or distant instances. Second, the Worker-Supervisor decomposition allows iterative correction of missed items and false positives. Third, AiRC adaptively stops refinement to balance accuracy and resource usage, avoiding premature stops or unnecessary iterations. We prioritize segmentation accuracy over latency, so the loop architecture may add VLM calls, but the controller bounds iterations to keep cost reasonable (2.6 iters on average, $\approx$\$0.0088 per sample with Gemini-2.5-flash; See supplementary). Finally, keeping VLM and SAM frozen avoids dataset‑specific overfitting, consistent with the gains we observe on ReasonSeg. Together, these choices address the weaknesses of open‑vocabulary methods under complex policies, resulting in substantial accuracy improvements and stable performance even on long prompt lengths.

\begin{figure}[t!]
  \centering
  \includegraphics[width=0.9\columnwidth]{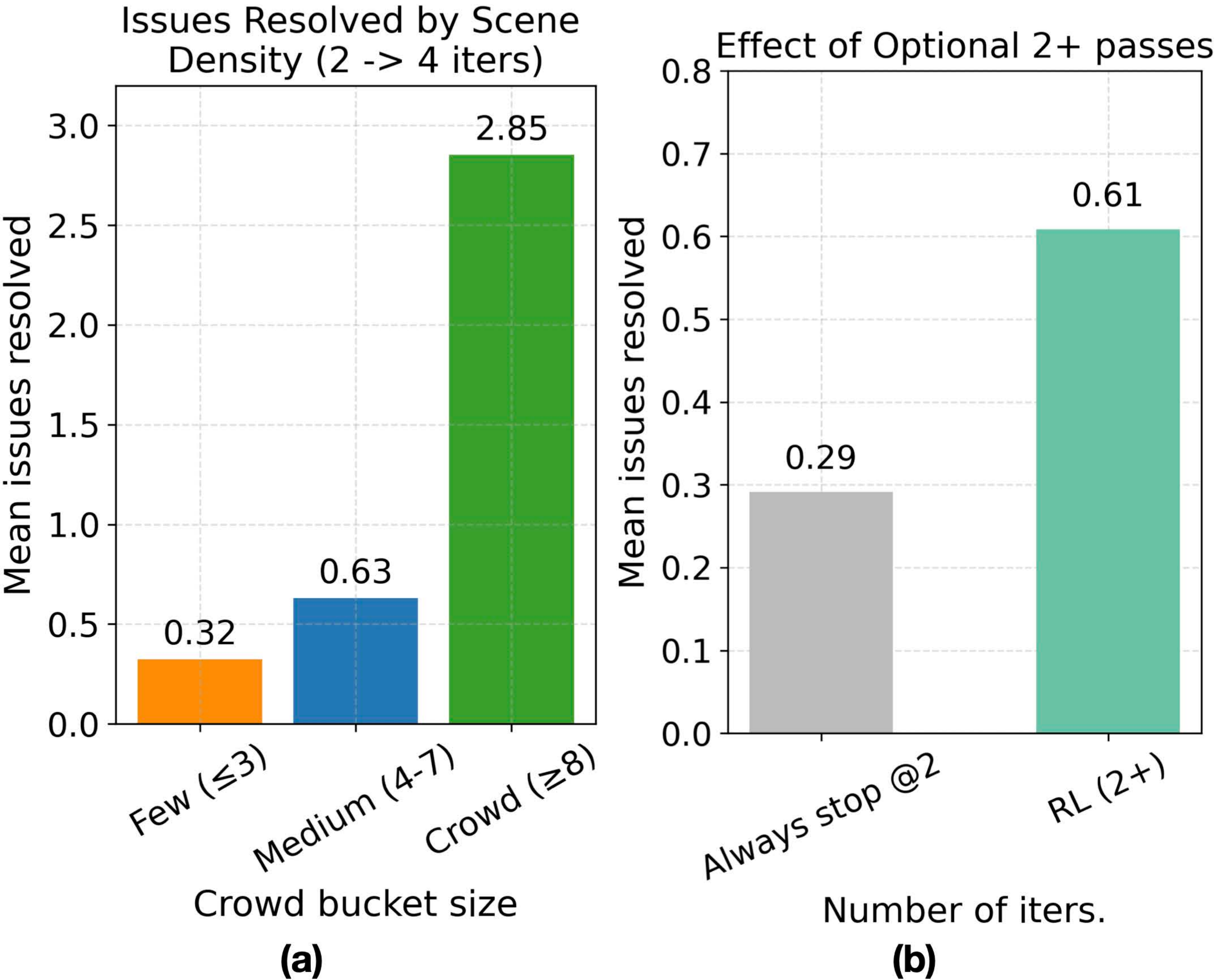}
  \caption{Effect of using RL controller measured in mean issues resolved over the dataset by (a) scene density and (b) letting AiRC dynamically make the stop decision and not hard-stopping at 2 iterations.}
  \label{fig:RLablation}
\end{figure}

\subsection{Ablation}
We ablate the iterative Worker–Supervisor loop by comparing pure Worker-only VLM detection against our context-enhanced and Supervisor-refined approaches (Table \ref{tab:ablation}). Providing targeted context boosts performance, with further improvement from Supervisor feedback. On Waymo, increases in IoU and recall indicate effective recovery of missed objects while the precision gains reflect pruning of false positives (Fig. \ref{fig:ablation}-a). For ReasonSeg, context construction is unnecessary due to shorter prompts; however, Supervisor refinement still notably improves segmentation (Fig. \ref{fig:ablation}-b). These results confirm that iterative refinement substantially enhances guideline adherence compared to one-shot VLM inference.

We also evaluate the advantage of using an adaptive iteration controller instead of a fixed stop heuristic. Fig.\ref{fig:RLablation}-a shows the gains when we do not hard‑stop at \texttt{MIN\_ITERS}=2. The largest gains on crowded scenes indicate that extra effort is directed where violations are most likely. Fig.\ref{fig:RLablation}-b shows that, on average, our AiRC module resolves 0.61 violations per crop versus 0.29 for a hard stop at two passes, a 110\% improvement at the cost of running extra passes on 48\% of crops. Both plots effectively show that dynamic stopping allocates compute to difficult scenes and increases overall quality at modest additional cost.

\subsection{Limitations and Future Work}
Our approach depends on clear, comprehensive guidelines; ambiguity or conflicts in the instructions can degrade its stability. Also, our current system's dependency on a proprietary VLM API constrains portability and on‑premises deployment. This was a necessary trade-off, as the Supervisor's critiquing role demands a level of long-context reasoning and instruction adherence that open-weight models do not yet reliably offer. We also prioritize accuracy over latency, so multiple API calls limit real‑time use. In future work, we will extend the framework to instance segmentation alongside semantic masks, enabling explicit association of missed/false objects with the correct subject instances. We also plan to reduce the dependency on proprietary VLM APIs by evaluating open weight, locally deployable models such as LLaVA-NeXT \cite{liu2024llavanext} and Qwen2.5-VL \cite{Qwen2.5-VL} as they become increasingly capable. 

\nocite{medeiros2023langsegmentanything}

\section{Conclusion}
We introduce guideline-consistent semantic segmentation, emphasizing strict adherence to detailed textual guidelines crucial in real-world applications. Our novel, training-free framework employs a multi-agent Worker-Supervisor architecture enhanced by a lightweight reinforcement learning stop policy, iteratively refines segmentation outputs to ensure compliance with complex, paragraph-length instructions without retraining. Evaluations on the Waymo and ReasonSeg datasets demonstrate that our approach notably surpasses existing state-of-the-art methods, highlighting its robustness, generalization capability, and efficiency in handling intricate guideline sets. 

\bibliography{aaai2026}

@misc{ren2024grounded,
      title={Grounded SAM: Assembling Open-World Models for Diverse Visual Tasks}, 
      author={Tianhe Ren and Shilong Liu and Ailing Zeng and Jing Lin and Kunchang Li and He Cao and Jiayu Chen and Xinyu Huang and Yukang Chen and Feng Yan and Zhaoyang Zeng and Hao Zhang and Feng Li and Jie Yang and Hongyang Li and Qing Jiang and Lei Zhang},
      year={2024},
      eprint={2401.14159},
      archivePrefix={arXiv},
      primaryClass={cs.CV}
}

@article{Qwen2.5-VL,
  title={Qwen2.5-VL Technical Report},
  author={Bai, Shuai and Chen, Keqin and Liu, Xuejing and Wang, Jialin and Ge, Wenbin and Song, Sibo and Dang, Kai and Wang, Peng and Wang, Shijie and Tang, Jun and Zhong, Humen and Zhu, Yuanzhi and Yang, Mingkun and Li, Zhaohai and Wan, Jianqiang and Wang, Pengfei and Ding, Wei and Fu, Zheren and Xu, Yiheng and Ye, Jiabo and Zhang, Xi and Xie, Tianbao and Cheng, Zesen and Zhang, Hang and Yang, Zhibo and Xu, Haiyang and Lin, Junyang},
  journal={arXiv preprint arXiv:2502.13923},
  year={2025}
}

@misc{liu2024llavanext,
    title={LLaVA-NeXT: Improved reasoning, OCR, and world knowledge},
    url={https://llava-vl.github.io/blog/2024-01-30-llava-next/},
    author={Liu, Haotian and Li, Chunyuan and Li, Yuheng and Li, Bo and Zhang, Yuanhan and Shen, Sheng and Lee, Yong Jae},
    month={January},
    year={2024}
}

@INPROCEEDINGS{giou_ciou,
  author={Mao, Junhua and Huang, Jonathan and Toshev, Alexander and Camburu, Oana and Yuille, Alan and Murphy, Kevin},
  booktitle={2016 IEEE Conference on Computer Vision and Pattern Recognition (CVPR)}, 
  title={Generation and Comprehension of Unambiguous Object Descriptions}, 
  year={2016},
  volume={},
  number={},
  pages={11-20},
  doi={10.1109/CVPR.2016.9}}

@inproceedings{kazemzadeh-etal-2014-referitgame,
    title = "{R}efer{I}t{G}ame: Referring to Objects in Photographs of Natural Scenes",
    author = "Kazemzadeh, Sahar  and
      Ordonez, Vicente  and
      Matten, Mark  and
      Berg, Tamara",
    editor = "Moschitti, Alessandro  and
      Pang, Bo  and
      Daelemans, Walter",
    booktitle = "Proceedings of the 2014 Conference on Empirical Methods in Natural Language Processing",
    month = oct,
    year = "2014",
    publisher = "Association for Computational Linguistics",
    url = "https://aclanthology.org/D14-1086/",
    doi = "10.3115/v1/D14-1086",
    pages = "787--798"
}

@article{Demir2018DeepGlobe2A,
  title={DeepGlobe 2018: A Challenge to Parse the Earth through Satellite Images},
  author={Ilke Demir and Krzysztof Koperski and David Lindenbaum and Guan Pang and Jing Huang and Saikat Basu and Forest Hughes and Devis Tuia and Ramesh Raskar},
  journal={2018 IEEE/CVF Conference on Computer Vision and Pattern Recognition Workshops},
  year={2018},
  pages={172-17209},
  url={https://api.semanticscholar.org/CorpusID:21727516}
}

@inproceedings{lisa,
  title={LISA: Reasoning Segmentation via Large Language Model},
  author={Lai, Xin and Tian, Zhuotao and Chen, Yukang and Li, Yanwei and Yuan, Yuhui and Liu, Shu and Jia, Jiaya},
  booktitle={Proceedings of the IEEE/CVF Conference on Computer Vision and Pattern Recognition},
  pages={9579-9589},
  year={2024}
}

@article{RL,
author = {Watkins, Christopher and Dayan, Peter},
year = {1992},
month = {05},
pages = {279-292},
title = {Technical Note: Q-Learning},
volume = {8},
journal = {Machine Learning},
doi = {10.1007/BF00992698}
}

@inproceedings{zhai2023sigmoid,
  title={Sigmoid loss for language image pre-training},
  author={Zhai, Xiaohua and Mustafa, Basil and Kolesnikov, Alexander and Beyer, Lucas},
  booktitle={Proceedings of the IEEE/CVF ICCV},
  pages={11975--11986},
  year={2023}
}

@InProceedings{waymo, author = {Sun, Pei and Kretzschmar, Henrik and Dotiwalla, Xerxes and Chouard, Aurelien and Patnaik, Vijaysai and Tsui, Paul and Guo, James and Zhou, Yin and Chai, Yuning and Caine, Benjamin and Vasudevan, Vijay and Han, Wei and Ngiam, Jiquan and Zhao, Hang and Timofeev, Aleksei and Ettinger, Scott and Krivokon, Maxim and Gao, Amy and Joshi, Aditya and Zhang, Yu and Shlens, Jonathon and Chen, Zhifeng and Anguelov, Dragomir}, title = {Scalability in Perception for Autonomous Driving: Waymo Open Dataset}, booktitle = {Proceedings of the IEEE/CVF CVPR}, month = {June}, year = {2020} }

@article{johnson2019billion,
  title={Billion-scale similarity search with {GPUs}},
  author={Johnson, Jeff and Douze, Matthijs and J{\'e}gou, Herv{\'e}},
  journal={IEEE Transactions on Big Data},
  volume={7},
  number={3},
  pages={535--547},
  year={2019},
  publisher={IEEE}
}

@inproceedings{reimers-2020-multilingual-sentence-bert,
  title     = {Making Monolingual Sentence Embeddings Multilingual using Knowledge Distillation},
  author    = {Reimers, Nils and Gurevych, Iryna},
  booktitle = {Proceedings of EMNLP},
  year      = {2020},
  publisher = {Association for Computational Linguistics},
  url       = {https://aclanthology.org/2020.emnlp-main.733},
  pages     = {849--862}
}

@article{team2025gemma,
  title={Gemma 3 technical report},
  author={Team Gemma and Kamath, Aishwarya and Ferret, Johan and Pathak, Shreya and Vieillard, Nino and Merhej, Ramona and Perrin, Sarah and Matejovicova, Tatiana and Ram{\'e}, Alexandre and Rivi{\`e}re, Morgane and others},
  journal={arXiv preprint arXiv:2503.19786},
  year={2025}
}

@article{minderer2023scaling,
  title={Scaling open-vocabulary object detection},
  author={Minderer, Matthias and Gritsenko, Alexey and Houlsby, Neil},
  journal={Advances in Neural Information Processing Systems},
  volume={36},
  pages={72983--73007},
  year={2023}
}

@inproceedings{
sun2025visual,
title={Visual Agents as Fast and Slow Thinkers},
author={Guangyan Sun and Mingyu Jin and Zhenting Wang and Cheng-Long Wang and Siqi Ma and Qifan Wang and Tong Geng and Ying Nian Wu and Yongfeng Zhang and Dongfang Liu},
booktitle={The Thirteenth International Conference on Learning Representations},
year={2025},
url={https://openreview.net/forum?id=ncCuiD3KJQ}
}

@article{kirillov2023segany,
  title={Segment Anything}, 
  author={Kirillov, Alexander and Mintun, Eric and Ravi, Nikhila and Mao, Hanzi and Rolland, Chloe and Gustafson, Laura and Xiao, Tete and Whitehead, Spencer and Berg, Alexander C. and Lo, Wan-Yen and Doll{\'a}r, Piotr and Girshick, Ross},
  journal={arXiv:2304.02643},
  year={2023}
}

@inproceedings{seem,
author = {Zou, Xueyan and Yang, Jianwei and Zhang, Hao and Li, Feng and Li, Linjie and Wang, Jianfeng and Wang, Lijuan and Gao, Jianfeng and Lee, Yong Jae},
title = {Segment everything everywhere all at once},
year = {2023},
publisher = {Curran Associates Inc.},
address = {Red Hook, NY, USA},
booktitle = {Proceedings of NeurIPS},
articleno = {868},
numpages = {14},
location = {New Orleans, LA, USA}
}

@inproceedings{radford2021learning,
  title={Learning transferable visual models from natural language supervision},
  author={Radford, Alec and Kim, Jong Wook and Hallacy, Chris and Ramesh, Aditya and Goh, Gabriel and Agarwal, Sandhini and Sastry, Girish and Askell, Amanda and Mishkin, Pamela and Clark, Jack and others},
  booktitle={ICML},
  pages={8748--8763},
  year={2021},
  organization={PmLR}
}

@inproceedings{li2022blip,
  title={Blip: Bootstrapping language-image pre-training for unified vision-language understanding and generation},
  author={Li, Junnan and Li, Dongxu and Xiong, Caiming and Hoi, Steven},
  booktitle={International conference on machine learning},
  pages={12888--12900},
  year={2022},
  organization={PMLR}
}

@inproceedings{liu2024grounding,
  title={Grounding dino: Marrying dino with grounded pre-training for open-set object detection},
  author={Liu, Shilong and Zeng, Zhaoyang and Ren, Tianhe and Li, Feng and Zhang, Hao and Yang, Jie and Jiang, Qing and Li, Chunyuan and Yang, Jianwei and Su, Hang and others},
  booktitle={ECCV},
  pages={38--55},
  year={2024},
  organization={Springer}
}

@inproceedings{Cheng2024YOLOWorld,
  title={YOLO-World: Real-Time Open-Vocabulary Object Detection},
  author={Cheng, Tianheng and Song, Lin and Ge, Yixiao and Liu, Wenyu and Wang, Xinggang and Shan, Ying},
  booktitle={Proc. IEEE Conf. Computer Vision and Pattern Recognition (CVPR)},
  year={2024}
}

@inproceedings{liang2023open,
  title={Open-vocabulary semantic segmentation with mask-adapted {CLIP}},
  author={Liang, Feng and Wu, Bichen and Dai, Xiaoliang and Li, Kunpeng and Zhao, Yinan and Zhang, Hang and Zhang, Peizhao and Vajda, Peter and Marculescu, Diana},
  booktitle={Proceedings of the IEEE/CVF CVPR},
  pages={7061--7070},
  year={2023}
}

@article{liu2023llava,
  title={Visual instruction tuning},
  author={Liu, Haotian and Li, Chunyuan and Wu, Qingyang and Lee, Yong Jae},
  journal={NeurIPS},
  volume={36},
  pages={34892--34916},
  year={2023}
}

@inproceedings{xia2024gsva,
  title={{GSVA}: Generalized segmentation via multimodal large language models},
  author={Xia, Zhuofan and Han, Dongchen and Han, Yizeng and Pan, Xuran and Song, Shiji and Huang, Gao},
  booktitle={Proceedings of the IEEE/CVF CVPR},
  pages={3858--3869},
  year={2024}
}

@article{comanici2025gemini,
  title={Gemini 2.5: Pushing the frontier with advanced reasoning, multimodality, long context, and next generation agentic capabilities},
  author={Comanici, Gheorghe and Bieber, Eric and Schaekermann, Mike and Pasupat, Ice and Sachdeva, Noveen and Dhillon, Inderjit and Blistein, Marcel and Ram, Ori and Zhang, Dan and Rosen, Evan and others},
  journal={arXiv preprint arXiv:2507.06261},
  year={2025}
}

@misc{openai2024gpt4v,
  author       = {OpenAI},
  title        = {GPT-4V(ision) System Card},
  year         = {2023},
  howpublished = {\url{https://openai.com/index/gpt-4v-system-card/}},
  note         = {Accessed: 2025-11-15}
}

@article{wu2023visual,
  title={Visual chatgpt: Talking, drawing and editing with visual foundation models},
  author={Wu, Chenfei and Yin, Shengming and Qi, Weizhen and Wang, Xiaodong and Tang, Zecheng and Duan, Nan},
  journal={arXiv preprint arXiv:2303.04671},
  year={2023}
}

@article{shen2023hugginggpt,
  title={Hugginggpt: Solving ai tasks with chatgpt and its friends in hugging face},
  author={Shen, Yongliang and Song, Kaitao and Tan, Xu and Li, Dongsheng and Lu, Weiming and Zhuang, Yueting},
  journal={Advances in Neural Information Processing Systems},
  volume={36},
  pages={38154--38180},
  year={2023}
}

@inproceedings{self_refine,
author = {Madaan, Aman and Tandon, Niket and Gupta, Prakhar and Hallinan, Skyler and Gao, Luyu and Wiegreffe, Sarah and Alon, Uri and Dziri, Nouha and Prabhumoye, Shrimai and Yang, Yiming and Gupta, Shashank and Majumder, Bodhisattwa Prasad and Hermann, Katherine and Welleck, Sean and Yazdanbakhsh, Amir and Clark, Peter},
title = {SELF-REFINE: iterative refinement with self-feedback},
year = {2023},
publisher = {Curran Associates Inc.},
address = {Red Hook, NY, USA},
booktitle = {Proceedings of the 37th International Conference on NeurIPS},
articleno = {2019},
numpages = {61},
location = {New Orleans, LA, USA},
}

@inproceedings{reflexion,
author = {Shinn, Noah and Cassano, Federico and Gopinath, Ashwin and Narasimhan, Karthik and Yao, Shunyu},
title = {Reflexion: language agents with verbal reinforcement learning},
year = {2023},
publisher = {Curran Associates Inc.},
address = {Red Hook, NY, USA},
booktitle = {Proceedings of the NeurIPS},
articleno = {377},
numpages = {19},
location = {New Orleans, LA, USA},
}

@inproceedings{Liao2025SelfCorrection,
  title     = {Can Large Vision–Language Models Correct Semantic Grounding Errors By Themselves?},
  author    = {Yuan-Hong Liao and Rafid Mahmood and Sanja Fidler and David Acuna},
  booktitle = {Proceedings of the IEEE/CVF Conference on Computer Vision and Pattern Recognition (CVPR)},
  year      = {2025},
  url       = {https://openaccess.thecvf.com/content/CVPR2025/papers/Liao_Can_Large_Vision-Language_Models_Correct_Semantic_Grounding_Errors_By_Themselves_CVPR_2025_paper.pdf}
}

@article{yang2023set,
  title={Set-of-mark prompting unleashes extraordinary visual grounding in gpt-4v},
  author={Yang, Jianwei and Zhang, Hao and Li, Feng and Zou, Xueyan and Li, Chunyuan and Gao, Jianfeng},
  journal={arXiv preprint arXiv:2310.11441},
  year={2023}
}

@INPROCEEDINGS{Sun2021IterativeShrinking,
  author={Sun, Mingjie and Xiao, Jimin and Lim, Eng Gee},
  booktitle={2021 IEEE/CVF Conference on Computer Vision and Pattern Recognition (CVPR)}, 
  title={Iterative Shrinking for Referring Expression Grounding Using Deep Reinforcement Learning}, 
  year={2021},
  volume={},
  number={},
  pages={14055-14064},
  doi={10.1109/CVPR46437.2021.01384}}

@INPROCEEDINGS {Su2024ScanFormer,
author = { Su, Wei and Miao, Peihan and Dou, Huanzhang and Li, Xi },
booktitle = { 2024 IEEE/CVF CVPR },
title = {{ ScanFormer: Referring Expression Comprehension by Iteratively Scanning }},
year = {2024},
volume = {},
ISSN = {},
pages = {13449-13458},
doi = {10.1109/CVPR52733.2024.01277},
url = {https://doi.ieeecomputersociety.org/10.1109/CVPR52733.2024.01277},
publisher = {IEEE Computer Society},
address = {Los Alamitos, CA, USA},
month =Jun}

@inproceedings{Wan2024CRG,
author = {Wan, David and Cho, Jaemin and Stengel-Eskin, Elias and Bansal, Mohit},
title = {Contrastive Region Guidance: Improving Grounding in Vision-Language Models Without Training},
year = {2024},
isbn = {978-3-031-72985-0},
publisher = {Springer-Verlag},
address = {Berlin, Heidelberg},
url = {https://doi.org/10.1007/978-3-031-72986-7_12},
doi = {10.1007/978-3-031-72986-7_12},
booktitle = {Computer Vision – ECCV 2024: 18th European Conference, Milan, Italy, September 29–October 4, 2024, Proceedings, Part LXXIX},
pages = {198–215},
numpages = {18},
location = {Milan, Italy}
}

@article{wang2022medical,
  title={Medical image segmentation using deep learning: A survey},
  author={Wang, Risheng and Lei, Tao and Cui, Ruixia and Zhang, Bingtao and Meng, Hongying and Nandi, Asoke K},
  journal={IET image processing},
  volume={16},
  number={5},
  pages={1243--1267},
  year={2022},
  publisher={Wiley Online Library}
}

@article{elhassan2024realtimeseg_ad_survey,
  title   = {Real-time Semantic Segmentation for Autonomous Driving: A Review of CNNs, Transformers, and Beyond},
  author  = {Elhassan, Mohammed A. M. and Zhou, Changjun},
  journal = {Journal of King Saud University - Computer and Information Sciences},
  year    = {2024},
  doi     = {10.1016/j.jksuci.2024.102226},
  url     = {https://www.sciencedirect.com/science/article/pii/S131915782400315X}
}

@article{al2024review,
  title={A review of visual SLAM for robotics: Evolution, properties, and future applications},
  author={Al-Tawil, Basheer and Hempel, Thorsten and Abdelrahman, Ahmed and Al-Hamadi, Ayoub},
  journal={Frontiers in Robotics and AI},
  volume={11},
  pages={1347985},
  year={2024},
  publisher={Frontiers Media SA}
}

@inproceedings{ronneberger2015u,
  title        = {{U-Net}: Convolutional Networks for Biomedical Image Segmentation},
  author       = {Ronneberger, Olaf and Fischer, Philipp and Brox, Thomas},
  booktitle    = {Medical Image Computing and Computer-Assisted Intervention – MICCAI 2015},
  volume       = {9351},
  pages        = {234--241},
  publisher    = {Springer},
  year         = {2015}
}

@inproceedings{chen2018encoder,
  title        = {Encoder-Decoder with Atrous Separable Convolution for Semantic Image Segmentation},
  author       = {Chen, Liang-Chieh and Zhu, Yukun and Papandreou, George and Schroff, Florian and Adam, Hartwig},
  booktitle    = {European Conference on Computer Vision (ECCV)},
  pages        = {833--851},
  year         = {2018},
  publisher    = {Springer}
}

@inproceedings{wu2024autogen,
  title={Autogen: Enabling next-gen LLM applications via multi-agent conversations},
  author={Wu, Qingyun and Bansal, Gagan and Zhang, Jieyu and Wu, Yiran and Li, Beibin and Zhu, Erkang and Jiang, Li and Zhang, Xiaoyun and Zhang, Shaokun and Liu, Jiale and others},
  booktitle={First Conference on Language Modeling},
  year={2024}
}

@inproceedings{hong2023metagpt,
  title={MetaGPT: Meta programming for a multi-agent collaborative framework},
  author={Hong, Sirui and Zhuge, Mingchen and Chen, Jonathan and Zheng, Xiawu and Cheng, Yuheng and Wang, Jinlin and Zhang, Ceyao and Wang, Zili and Yau, Steven Ka Shing and Lin, Zijuan and others},
  booktitle={The Twelfth International Conference on Learning Representations},
  year={2023}
}

@article{hurst2024gpt,
  title={Gpt-4o system card},
  author={Hurst, Aaron and Lerer, Adam and Goucher, Adam P and Perelman, Adam and Ramesh, Aditya and Clark, Aidan and Ostrow, AJ and Welihinda, Akila and Hayes, Alan and Radford, Alec and others},
  journal={arXiv preprint arXiv:2410.21276},
  year={2024}
}

@misc{medeiros2023langsegmentanything,
  author       = {Luca Medeiros},
  title        = {{Language Segment‑Anything}},
  howpublished = {\url{https://github.com/luca-medeiros/lang-segment-anything}},
  year         = {2023},
}

@article{segzero,
  title={Seg-zero: Reasoning-chain guided segmentation via cognitive reinforcement},
  author={Liu, Yuqi and Peng, Bohao and Zhong, Zhisheng and Yue, Zihao and Lu, Fanbin and Yu, Bei and Jia, Jiaya},
  journal={arXiv preprint arXiv:2503.06520},
  year={2025}
}

@inproceedings{read,
  title={Reasoning to attend: Try to understand how {SEG} token works},
  author={Qian, Rui and Yin, Xin and Dou, Dejing and Dou, Dejing},
  booktitle={Proceedings of the CVPR Conference},
  pages={24722--24731},
  year={2025}
}

@article{ravi2024sam2,
  title={SAM 2: Segment Anything in Images and Videos},
  author={Ravi, Nikhila and Gabeur, Valentin and Hu, Yuan-Ting and Hu, Ronghang and Ryali, Chaitanya and Ma, Tengyu and Khedr, Haitham and R{\"a}dle, Roman and Rolland, Chloe and Gustafson, Laura and Mintun, Eric and Pan, Junting and Alwala, Kalyan Vasudev and Carion, Nicolas and Wu, Chao-Yuan and Girshick, Ross and Doll{\'a}r, Piotr and Feichtenhofer, Christoph},
  journal={arXiv preprint arXiv:2408.00714},
  url={https://arxiv.org/abs/2408.00714},
  year={2024}
}
\appendix
\clearpage

\twocolumn[
  {%
    \centering
    \bfseries\huge Supplementary Material\\[0.5em]\LARGE Guideline‑Consistent Segmentation via Multi‑Agent Refinement\\[1em]  
    \addcontentsline{toc}{section}{Supplementary Material}%
    \vspace{1em}                                 
  }%
]

\setcounter{section}{0}
\renewcommand{\thesection}{\Alph{section}}

\setcounter{table}{0}
\renewcommand{\thetable}{\Alph{table}}

\setcounter{figure}{0}
\renewcommand{\thefigure}{\Alph{figure}}

\section{Contents}
The supplementary material is organized as follows:
\begin{itemize}
    \item Sec. A \ref{app:waymo_guidelines}: Example of the Official Waymo Guidelines
    \item Sec. B \ref{app:Airc}: Intuitive form of AiRC
    \item Sec. C \ref{app:dataset}: Dataset Curation
    \item Sec. D \ref{app:reasonseg}: ReasonSeg Qualitative Results
    \item Sec. E \ref{app:failure}: Examples of Failure Cases
    \item Sec. F \ref{app:prompts}: Example of System Prompts  
    \item Sec. G \ref{app:costs}: Estimated Costs and Latency
\end{itemize}

\vspace{1.5em} 

\subsection{Sec A: Example of the Official Waymo Guidelines}
\label{app:waymo_guidelines}
We reference the official Waymo guidelines in this study. The guidelines can also be found on the Waymo GitHub page and are also provided with our attached code files. 
\begin{itemize}
\small
    \item \texttt{All objects that can be recognized as a pedestrian and are at least partially visible are labeled.}
    \item \texttt{If it is not possible to tell by looking at the camera image whether an object is a pedestrian, the object is not labeled.}
    \item \texttt{People who are walking or riding kick scooters (including electric kick scooters), segways, skateboards, etc. are labeled as pedestrians.}
    \item \texttt{People inside other vehicles are not labeled, except for people standing on the top of cars/trucks or standing on flatbeds of trucks.}
    \item \texttt{A person riding a bicycle is not labeled as a pedestrian, but labeled as a cyclist instead.}
    \item \texttt{Mannequins, statues, billboards, posters, or reflections of people are not labeled.}
    \item \texttt{Single label (including the pedestrian and additional objects) is created if the pedestrian is holding a small child or carrying small items (smaller than 2m in size such as umbrella or small handbag or a sign) being held.}
    \item \texttt{Single label (including the pedestrian and additional objects) is created if the pedestrian is  riding a kick scooter (including electric kick scooter), a segway, a skateboard, etc.}
    \item \texttt{If the pedestrian is carrying an object larger than 2m, or pushing a bike or shopping cart, the label does not include the additional object.}
    \item \texttt{If the pedestrian is pushing a stroller with a child in it, separate bounding boxes are created for the pedestrian and the child. The stroller is not included in the child box.}
    \item \texttt{If pedestrians overlap each other, they are labeled as separate objects.}
\end{itemize}


\subsection{Sec B: Intuitive form of Decision Inequality for AiRC}
\label{app:Airc}
At iteration $t$, issue count $I_t$ and state $s$, let
\begin{align}
  \mu_{\Delta} &= \mathbb{E}\bigl[I_t - I_{t+1}\mid s_t,\;\mathrm{CONTINUE}\bigr],\\
  \pi_{\mathrm{clear}} &= \Pr\bigl[I_{t+1}=0\mid s_t,\;\mathrm{CONTINUE}\bigr],\\
  V_{t+1} &= \mathbb{E}\bigl[\max_{a'}Q(s_{t+1},a')\bigr].
\end{align}  
where $\mu_{\Delta}$ is the expected issues fixed, $\pi_{\mathrm{clear}}$ is the probability next pass finishes, and $V_{t+1}$ is the future value which is terminal if at the \texttt{MAX\_ITERS}. Then the Q-values describing the expected total future reward we will collect given state $s$, action $a$, and a current policy ($c=$step cost, $b=$clear bonus, $\gamma=$discount factor, $p=$early stop penalty), can be approximated by:
\begin{equation}
\begin{aligned}
Q(s_t,\mathrm{CONTINUE}) &\approx \mu_{\Delta} - c + b\,\pi_{\mathrm{clear}} + \gamma\,V_{t+1},\\
Q(s_t,\mathrm{STOP}) &\approx 
\begin{cases}
0,   & I_t = 0,\\
-p, & I_t > 0.
\end{cases}
\end{aligned}
\end{equation}

\begin{figure}[ht]
  \centering
\includegraphics[width=\columnwidth]{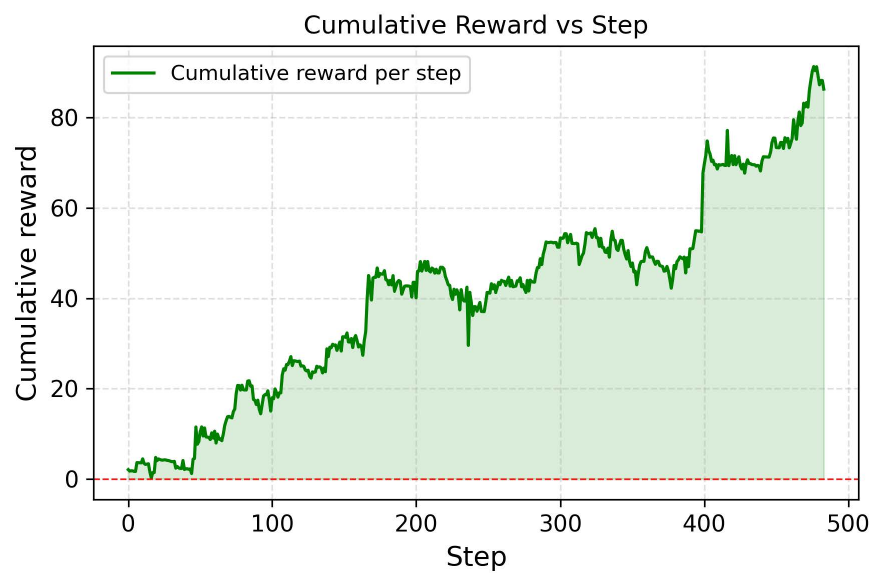}
  \caption{Cumulative reward progression demonstrates successful learning by the RL iteration controller optimizing its decision-making strategy.}
  \label{fig:RL_cumulative}
\end{figure}

Thus, AiRC chooses \textsc{CONTINUE} if 
\begin{equation}
    Q(s_t,\mathrm{CONTINUE}) > Q(s_t,\mathrm{STOP})
\end{equation}

\begin{figure*}[!ht]
  \centering
  \includegraphics[width=\linewidth]{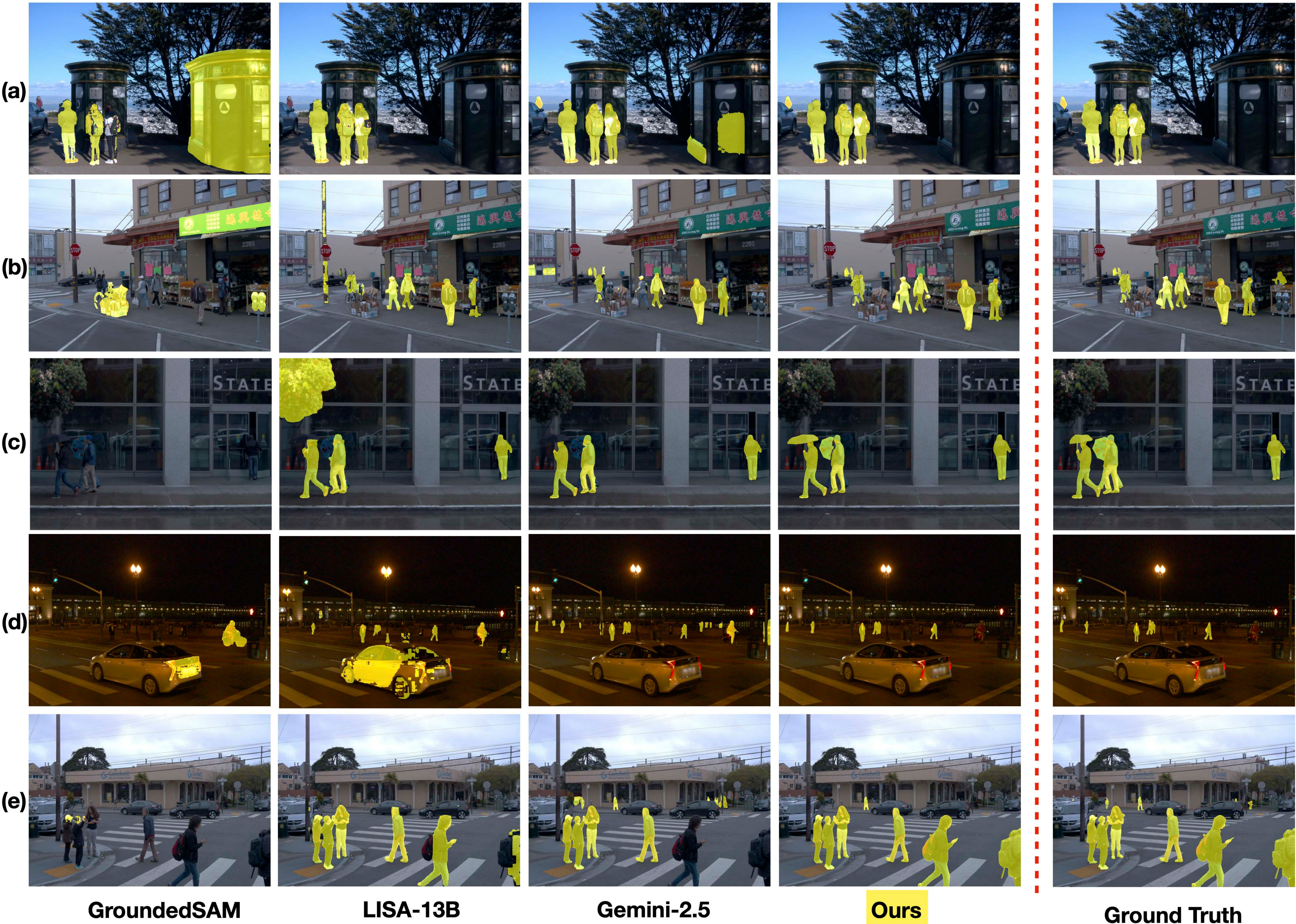}
  \caption{We curate a subset of the Waymo dataset which precisely follows the Waymo guideline manual. As seen in the figure, it contains diverse examples of scenes such as (a) sunny, (b)/(e) overcast, (c) rainy, and (d) night, with a mix of crowded and relatively non-crowded scenes. Our method performs consistently better than the SOTA and the baseline.}
  \label{fig:appendix_qual}
\end{figure*}

In other words, AiRC module chooses CONTINUE if its Q-value (expected discounted return) in the current state is higher than that of STOP; otherwise, it opts to stop. 

Cumulative reward vs. learning steps (Fig. \ref{fig:RL_cumulative}) demonstrates that our RL iteration controller effectively learns and improves performance over time. The trajectory never falls below zero, demonstrating that the agent's policy decisions consistently yield net positive outcomes. A negative cumulative reward would indicate that the agent's interventions were counterproductive, costing more than the value they provided.

\begin{figure*}[!t]
  \centering
  \includegraphics[width=\linewidth]{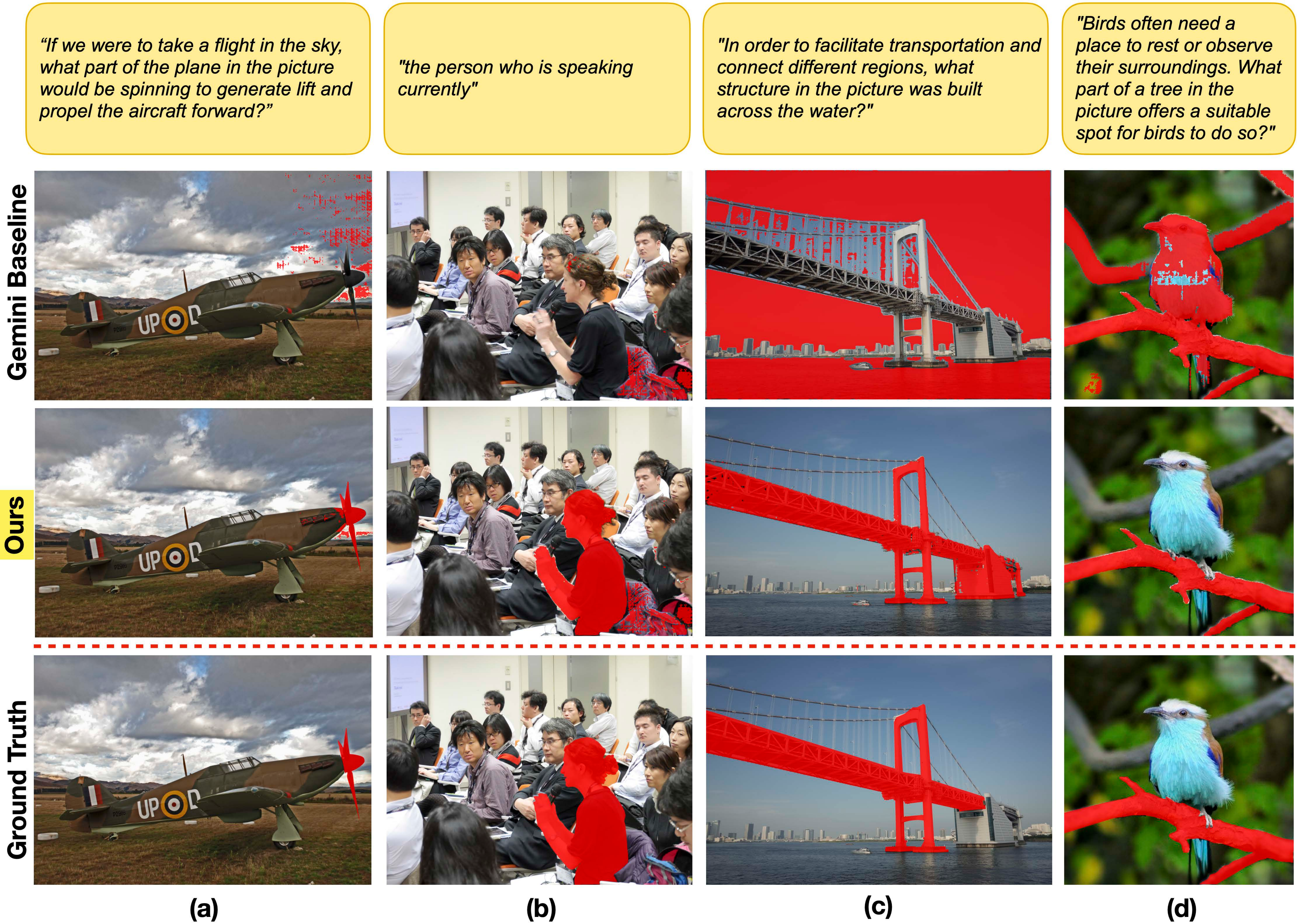}
  \caption{Qualitative results on the ReasonSeg dataset over our Gemini-2.5 baseline. The textboxes represent the input reasoning text.}
  \label{fig:reasonseg_qual}
\end{figure*}

\begin{figure*}[!t]
  \centering
  \includegraphics[width=0.9\linewidth]{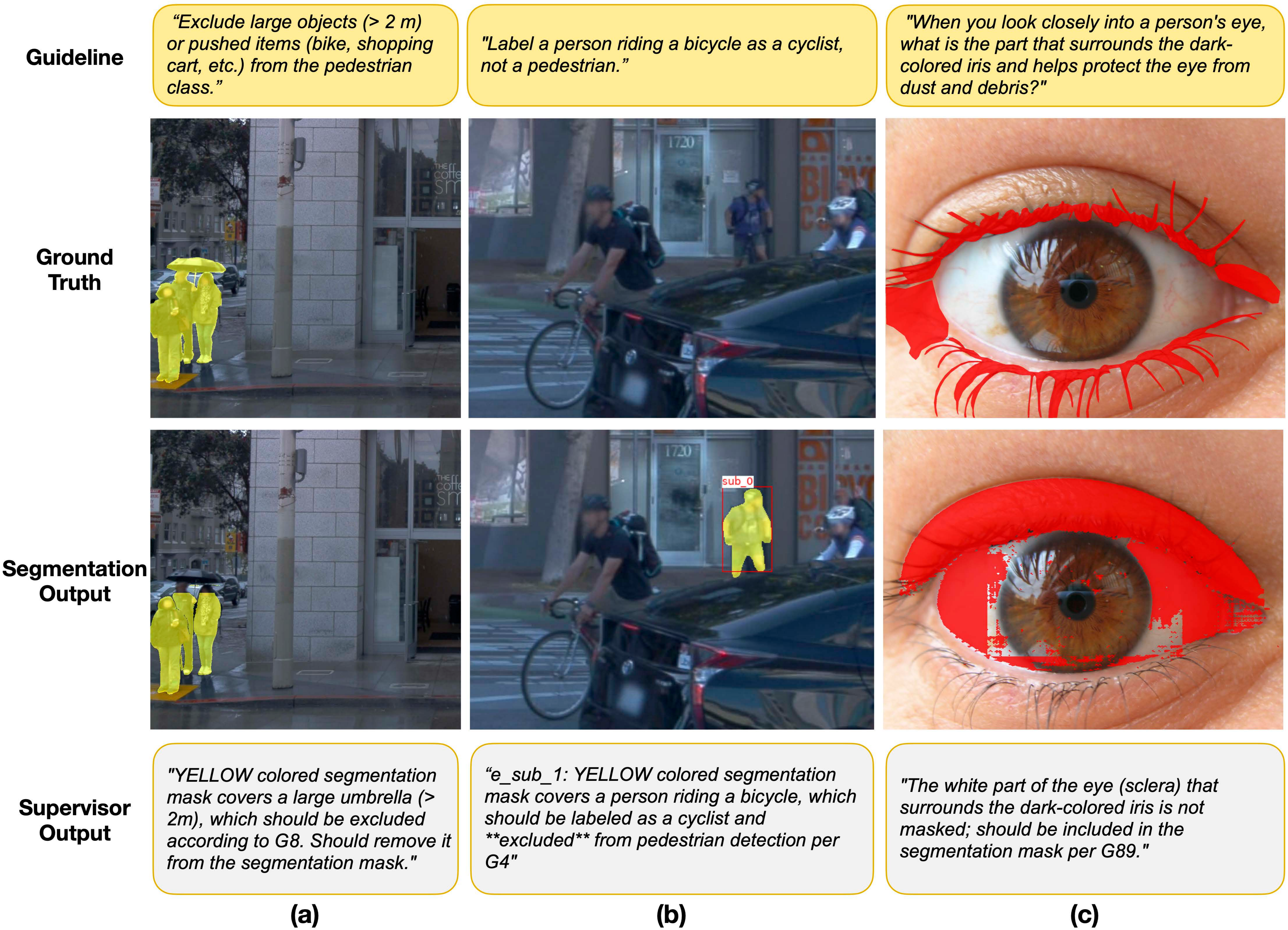}
  \caption{Failure cases on Waymo and ReasonSeg. (a) A handheld umbrella is wrongly excluded due to misinterpretation of size constraints. (b) One cyclist is corrected, but another remains mislabeled. (c) Fine-grained features like \textit{eye-lashes} are misclassified due to VLM errors and bounding box limitations.}
  \label{fig:failure_cases}
\end{figure*}

\begin{figure}[ht]
  \centering
  \includegraphics[width=\columnwidth]{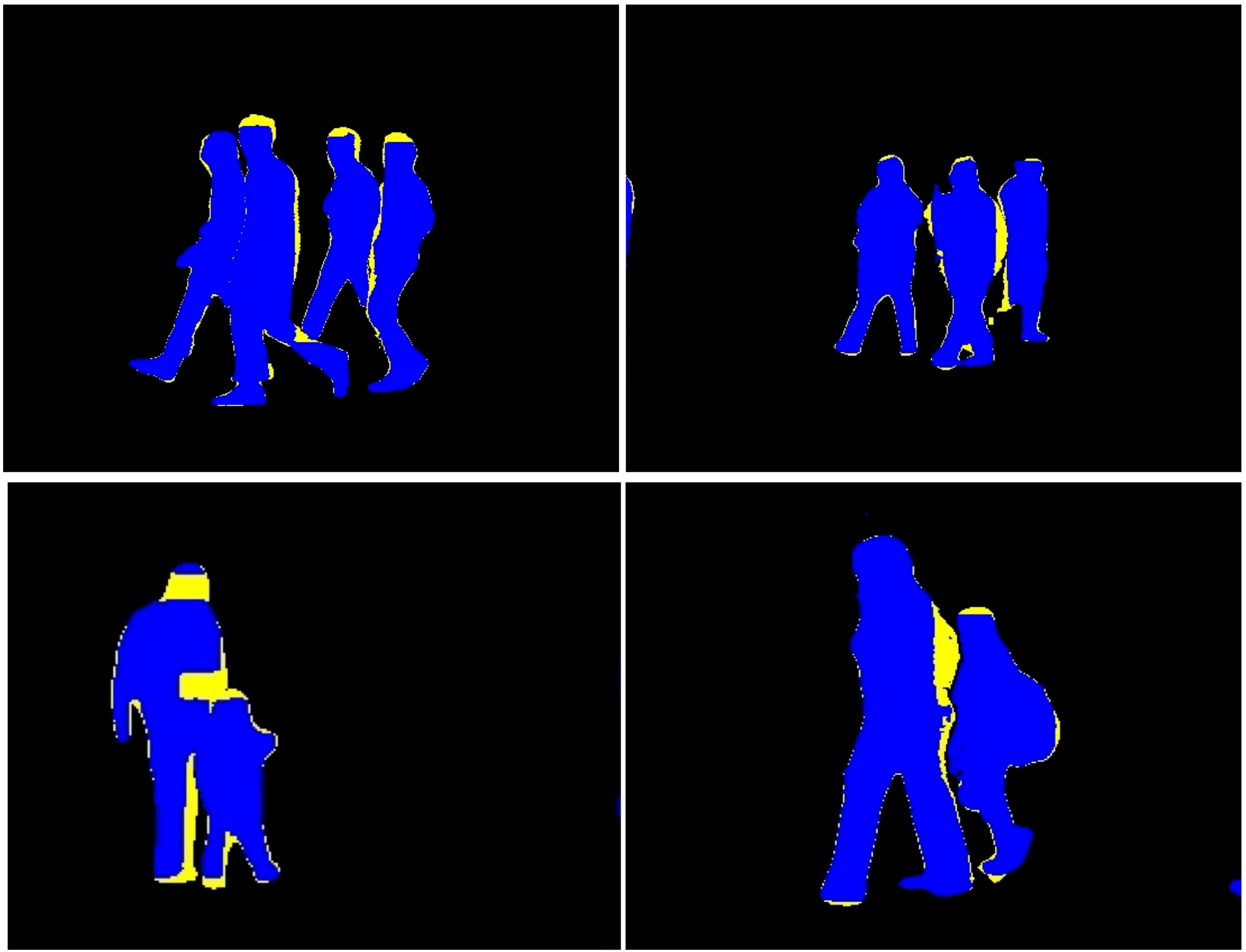}
  \caption{Effect of the Supervisor’s refinement step. Initial Worker masks (blue) are improved after fine-grained boundary corrections proposed by the Supervisor, with the resulting changes highlighted in yellow.}
  \label{fig:refinements}
\end{figure}


\subsection{Sec C: Dataset Curation}
\label{app:dataset}
Waymo labels often violate their own guidelines (Main text Fig.~\ref{fig:waymo_inconsistent}). As our method is designed to enforce strict guideline adherence, using noisy annotations could produce misleading outcomes. To ensure fair and accurate evaluation, we manually curate a set of 101 samples that strictly follow the official guidelines. This effort-intensive step was necessary to properly assess our core contribution. We first formed a candidate pool of 130 images that strictly satisfy the official Waymo guidelines with diverse weather conditions (sun/rain/night) and scene density. From this pool we randomly selected 101 images with k images per scene where $1\le k \le 7$. The 101 samples span 55 distinct scenes, providing broad coverage for guideline-consistent evaluation. Our results on the sampled subset show strong statistical significance over the Gemini baseline (paired t-test p$<$0.001, win-rate 75.8\%, and 95\%CI improvement=[+6.0,+13.8] gIoU), indicating that the gains are unlikely to be due to random variation even at this modest sample size. Fig. \ref{fig:appendix_qual} shows some examples.


\subsection{Sec D: ReasonSeg Qualitative Results}
\label{app:reasonseg}
Our method also performs well on reasoning segmentation dataset over the other SOTA (main text Table \ref{tab:reasonseg}). Here, we show the qualitative comparisons on ReasonSeg as compared to our baseline and SOTA Gemini-2.5 (Fig. \ref{fig:reasonseg_qual}).

\subsection{Sec E: Examples of Failure Cases}
\label{app:failure}
Although our method generally performs well, certain challenging scenarios reveal limitations. For instance, in the Waymo dataset (Fig. \ref{fig:failure_cases}-a), our framework mistakenly applies the guideline that excludes large objects (greater than 2m) to an umbrella, causing incorrect omission. This occurs because the framework struggles to precisely interpret quantitative constraints (such as size thresholds) and instead approximates object dimensions. Consequently, despite a competing guideline explicitly instructing inclusion of handheld items like umbrellas, the model overly prioritizes the size-related exclusion. In another Waymo case (Fig. \ref{fig:failure_cases}-b), while the Supervisor correctly follows the guideline of labeling a person riding a bicycle as a cyclist, promptly removing a misclassified pedestrian ($sub\_1$, not shown in the figure since it was removed), it fails to similarly recognize and reclassify another pedestrian instance ($sub\_0$) as a cyclist. Thus, the misclassification persists in the final segmentation. 
Similarly, on the ReasonSeg dataset (Fig. \ref{fig:failure_cases}-c), the VLM incorrectly outputs the answer as \textit{sclera} instead of \textit{eye-lashes}. Also, our approach predominantly relies on bounding box prompts to SAM, which can inherently limit fine-grained segmentation, making precise delineation of small or detailed features (like \textit{eye-lashes}) difficult.

To mitigate these issues, future enhancements might involve explicitly ranking guidelines to clarify priority conflicts and supplementing bounding boxes with precise point prompts to SAM, enabling finer segmentation accuracy. These refinements represent promising directions for improving guideline adherence and segmentation precision.

\lstdefinestyle{json}{
    basicstyle=\scriptsize\ttfamily,
    breaklines=true,
    frame=none,
    showstringspaces=false,
    captionpos=b,
    numbers=none
}

\begin{table*}[!ht]
\small
\centering
\begin{tabular}{|m{1.5cm}|m{2.8cm}|m{5.2cm}|m{5.2cm}|}
\hline
\textbf{Agent} & \textbf{Role} & \textbf{Example System Prompt} & \textbf{Expected Example Output} \\
\hline

\multirow{2}{*}{Worker} & Detection, Segmentation & 
\begin{lstlisting}[style=json, basicstyle=\scriptsize\ttfamily]
[Detect all objects 
from user prompt and the image. 
Return JSON]
\end{lstlisting} & 
\begin{lstlisting}[style=json, basicstyle=\scriptsize\ttfamily]
"instances": [{ 
  "id": "sub_0",
  "label": "pedestrian",
  "box_2d": [100, 100, 200, 200]
}]
\end{lstlisting} \\
\cline{2-4}

& Takes feedback from Supervisor & 
\begin{lstlisting}[style=json, basicstyle=\scriptsize\ttfamily]
[For each entry in 
'refinements', update the 
corresponding worker box via its 
'box_id']
\end{lstlisting} & 
\begin{lstlisting}[style=json, basicstyle=\scriptsize\ttfamily]
"instances": [{ 
  "id": "sub_0",
  "label": "pedestrian",
  "box_2d": [150, 150, 250, 250]
}]
\end{lstlisting} \\
\hline

Supervisor Eval & Critiques and proposed refinements & 
\begin{lstlisting}[style=json, basicstyle=\scriptsize\ttfamily]
[Output the following:
false_positives: Format: e_<n>,
label, reason, (cite G_<id>)
missing_objects: Format: m_<n>,
label, reason (cite G_<id>)
refinements: object already masked;
box edge needs minor nudge]
\end{lstlisting} & 
\begin{lstlisting}[style=json, basicstyle=\scriptsize\ttfamily]
"missing_objects": [{
  "missing_object_id": "m_1",
  "label": "umbrella", 
  "reason": "Umbrella should be 
  included per G<id>"
}],
"false_positives": [],
"refinements": []
\end{lstlisting} \\
\hline

Supervisor Boxgen & Generates candidate boxes & 
\begin{lstlisting}[style=json, basicstyle=\scriptsize\ttfamily]
[return box_2d for 
entry in 'missing_objects'/
'false_positives' per their label, 
id, and reason.]
\end{lstlisting} & 
\begin{lstlisting}[style=json, basicstyle=\scriptsize\ttfamily]
"instances": [{
  "box_id": "m_1",
  "label": "umbrella",
  "box_2d": [123, 456, 789, 987]
}]
\end{lstlisting} \\
\hline

\end{tabular}
\caption{Multi-agent configuration used in our critique-and-refine loop. Each row shows an agent’s role, an abbreviated system prompt, and a toy JSON response. Worker detects/segments objects and later applies the Supervisor’s edits. Supervisor-Eval returns three ordered lists: missing\_objects, false\_positives, and refinements, with their brief rationales. Supervisor-Boxgen proposes 2D boxes for each missing or false-positive item. Full prompt templates and JSON schemas are released with the code.}
\label{tab:appendix_prompts}
\end{table*}


\subsection{Sec F: Example of System Prompts}
\label{app:prompts}
We structure the VLM into specialized agents by assigning each a specific role and corresponding responsibilities via explicit system prompts. A concise summary is shown in Table~\ref{tab:appendix_prompts}, with full prompts available in the supplementary code files.

\subsubsection{Worker}
The Worker VLM agent performs the initial segmentation task based on textual instructions, outputting detected object labels and their bounding boxes ($box\_2d$ coordinates). The recommended way to get a quality structured output JSON out of Gemini-2.5 is to properly define a $response\_schema$, allowing precise extraction and standardization of outputs for subsequent processing. However, we observed that this constraint reduced its bounding-box detection quality. Thus, we allow the Worker to output detections freely, appending unique subject identifiers ($sub\_i$) to each detected instance. The bounding boxes are then passed to SAM for mask generation. The Worker’s final outputs, the JSON results ($worker\_output\_json$), annotated images ($worker\_annotated\_image$) with masks and bounding boxes labeled by the unique $sub\_i$, and original input image ($original\_image$), are sent to the Supervisor for evaluation.

\subsubsection{Supervisor.}
Since the $Agent\_1: supervisor\_eval$ only has the critiquing capability and not object detection, we can define a $response\_schema$ to get the exact structured output. Specifically, the Supervisor reviews the guidelines and identifies any missed objects ($missing\_objects$: $m\_1, m\_2, ..., m\_n$) and false positives ($false\_positives$: $e\_1, e\_2, ..., e\_n$), along with brief descriptions. For refinements, it refers explicitly to detected instances via their unique IDs ($sub\_i$) and suggests necessary adjustments (Fig. \ref{fig:refinements}). We keep dynamic thinking mode active only for the Supervisor Agent 1, enhancing guideline comprehension and decision-making. A structured JSON with missing/false objects with their labels and description, and the $original\_image$ goes to Agent 2. We also send $worker\_annotated\_image$ to Agent 2 so that it knows what is already masked.

$Agent\_2: supervisor\_boxgen$ is similar to the Worker agent, but it takes the description of each box from the supervisor to help it localize where the missing/false object can be. Since this is also object detection, we do not put the $response\_schema$ for higher quality object detection. 
The final structured JSON is returned as $missing\_objects$, $false\_positives$, and $refinements$, each with their $IDs$, $box\_2d$, $labels$ and a brief $description$.

A SigLIP verifier then filters the proposed candidates by computing a cosine similarity between image embeddings obtained from cropping the image to $box\_2d$, and text embeddings from $labels$. The candidates with similarity greater than or equal to 0.5 are kept, otherwise, those keys are deleted from the output JSON. 

This final JSON goes back to the worker, which it adds/removes the proposed candidates and updates the boxes for segmentation if any refinement is present.

\vspace{1.5em} 

\subsection{Sec G: Estimated Costs and Latency}
\label{app:costs}
Our framework prioritizes accuracy over latency, as segmentation quality is critical. Nonetheless, we ensure processing remains cost-effective. Using \texttt{Gemini-2.5-flash-preview-05-20} (priced at \$0.30 per million input tokens and \$2.50 per million output tokens), each Worker or Supervisor call consumes roughly 2000 input and 200 output tokens, costing approximately \$0.0011 per API call with a median latency of around 1.1s. A full Worker-Supervisor iteration involves three API calls, leading to per-sample costs ranging from \$0.0066 (minimum 2 iterations) to \$0.0132 (maximum 4 iterations). Crucially, our Adaptive Iteration Controller (AiRC) stops the loop once the improvement stagnates, empirically averaging around 2.66 iterations, thus keeping the typical cost closer to \$0.0088 per sample. Thus, segmenting 1000 samples costs between \$6.6-\$13.2 and \$8.8 on average, making large-scale, guideline-consistent segmentation economically feasible.

\end{document}